\documentclass[final,5p,times,twocolumn,numeric]{elsarticle}

\usepackage{makecell}
\usepackage{amssymb}
\usepackage{adjustbox}
\usepackage{lipsum}
\usepackage{enumitem}
\usepackage{tikz}
\usepackage{fontawesome}
\usepackage{scalerel}
\usepackage[edges]{forest}
\definecolor{hidden-draw}{RGB}{0,0,0}
\definecolor{hidden-pink}{RGB}{168,191,143}

\usepackage[colorlinks]{hyperref}
\usepackage{amsmath} 
\usepackage{multirow}
\usepackage{bbding}
\usepackage[T1]{fontenc}
\usepackage{graphicx}
\usepackage{subfig}

\usepackage{bm}
\usepackage{colortbl}  
\usepackage{array} 
\usepackage{tabularx}
\usepackage{framed}
\usepackage{ulem}
\usepackage{booktabs}
\usepackage{stfloats}
\usepackage{xcolor}
\usepackage{mydef}

\definecolor{mycolor}{RGB}{134,150,167}
\definecolor{myorange}{RGB}{253,229,206}
\definecolor{mygreen}{RGB}{213,232,212}
\definecolor{myblue}{RGB}{180,199,231}
\definecolor{slightblue}{RGB}{189,215,238}

\definecolor{reference}{RGB}{55,126,168}
\AtBeginDocument{
  \hypersetup{
            linkcolor=reference,
            anchorcolor=reference,
            citecolor=reference,
            filecolor=reference,
            urlcolor=reference
}
}

\journal{}

\begin{document}

\begin{frontmatter}



\title{Geospatial Representation Learning: A Survey \\from Deep Learning to The LLM Era}


\author[ustgz]{Xixuan Hao}
\author[ustgz]{Yutian Jiang}
\author[ustgz]{Xingchen Zou}
\author[ustgz]{Jiabo Liu}

\author[astar]{\\Yifang Yin}
\author[uwmadison]{Song Gao}
\author[unsw]{Flora Salim}
\author[swjtu]{Tianrui Li}
\author[ustgz]{Yuxuan Liang\corref{cor1}}
\affiliation[ustgz]{organization={The Hong Kong University of Science and Technology (Guangzhou)},
            city={Guangzhou},
            country={China}}
\affiliation[uwmadison]{organization={University of Wisconsin-Madison},
            city={Madison},
            country={USA}}
\affiliation[unsw]{organization={The University of New South Wales},
            city={Sydney},
            country={Australia}}
\affiliation[swjtu]{organization={Southwest Jiaotong University},
            city={Chengdu},
            country={China}}
\affiliation[astar]{organization={Institute for Infocomm Research (I$^2$R), A*STAR},
            country={Singapore}}
 \cortext[cor1]{Y. Liang is the corresponding author. Email: yuxliang@outlook.com}


\begin{abstract}

The ability to transform location-centric geospatial data into meaningful computational representations has become fundamental to modern spatial analysis and decision-making. 
Geospatial Representation Learning (GRL), the process of automatically extracting latent structures and semantic patterns from geographic data, is undergoing a profound transformation through two successive technological revolutions: the deep learning breakthrough and the emerging large language model (LLM) paradigm. While deep neural networks (DNNs) have demonstrated remarkable success in automated feature extraction from structured and semi-structured geospatial data (e.g., satellite imagery, GPS trajectories), the recent integration of LLMs introduces transformative capabilities for cross-modal geospatial reasoning and unstructured geo-textual data processing.
This survey presents a comprehensive review of geospatial representation learning across both technological eras, organizing them into a structured taxonomy based on the complete pipeline comprising: (1) data perspective, (2) methodological perspective, and (3) application perspective.
We also highlight current advancements, discuss existing limitations, and propose
potential future research directions in the LLM and foundation model era. This work offers a thorough
exploration of the field and provides a roadmap for further innovation in GRL. The summary of the up-to-date paper list can be found in \url{https://github.com/CityMind-Lab/Awesome-Geospatial-Representation-Learning} and will undergo continuous updates.
\end{abstract}




\end{frontmatter}





\section{Introduction}

In the era of ubiquitous geospatial sensing and AI proliferation, the ability to extract actionable insights from location-centric data has become a critical driver of scientific discovery and societal transformation. 
The exponential growth of multimodal geospatial data, from satellite imagery to human mobility and geo-textual records, has catalyzed a fundamental shift in how we understand and model geographic phenomena. 
Traditional Geographic Information Systems (GIS), while effective for basic spatial operations~\cite{bonham1994geographic,maguire1991overview}, are increasingly complemented by Geospatial AI (GeoAI) – a transformative fusion of deep learning and spatial science that unlocks unprecedented capabilities in modeling geographic complexity~\cite{gao2023handbook,helbich2019using,song2016deeptransport}.

At the heart of this emerging paradigm lies in \textbf{Geospatial Representation Learning (GRL)}. This discipline aims to extract latent structures, semantic patterns, and geographic dependencies from heterogeneous location-centric data by incorporating spatial relationships, temporal variations, and environmental contexts~\citep{mai2023spatial}. These learned representations support a wide range of geospatial analytical tasks, from prediction and classification to simulation and spatial reasoning, thereby strengthening both scientific inquiry and real-world decision-making~\citep{zou2025deep,ufmsurvey-kdd2024}. In essence, GRL serves as the conceptual and technical backbone of modern location-based systems and services (LBS).

\begin{figure*}[!t]
    \centering
    \includegraphics[width=0.99\linewidth]
    {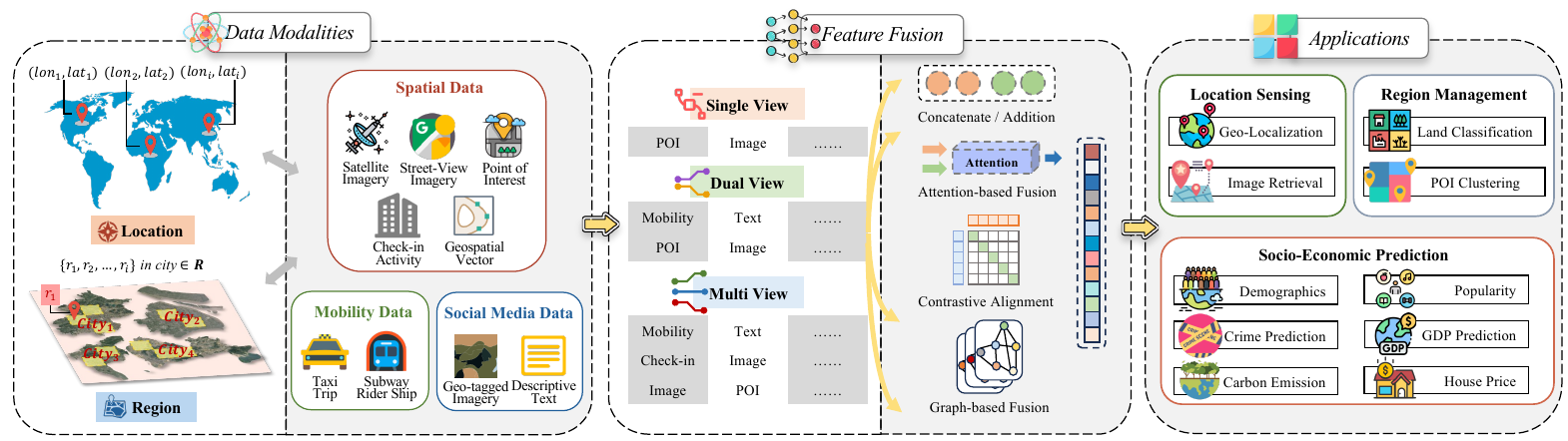} 
    \caption{The complete pipeline of geospatial representation learning with location-centric data.}
    \label{fig:intro}
\end{figure*}

The evolution of GRL has undergone two revolutionary phases. The first wave, driven by deep neural networks, established new benchmarks in spatial pattern recognition through architectures like Convolutional Neural Networks (CNNs) for satellite image analysis~\cite{lecun2015deep}, Graph Neural Networks (GNNs) for urban network modeling~\cite{MVURE}, and Recurrent neural networks (RNNs) for trajectory prediction~\cite{capobianco2021deep}. These approaches demonstrated remarkable success in automating feature extraction from structured and semi-structured geospatial data (location coordinates, vector geometries, raster images, trajectories), enabling applications ranging from climate change prediction~\cite{wang2024nuwadynamics} to intelligent transportation systems~\cite{yuan2021survey}. However, the emerging second wave, propelled by large language models (LLMs) and multimodal foundation models (FMs), is redefining the boundaries of GRL~\cite{mai2024opportunities}. Modern GeoAI systems now integrate unstructured textual data (social media geo-tags, administrative reports) with conventional spatial data streams through cross-modal alignment~\cite{zou2025deep}, while leveraging pretrained knowledge from LLMs to enhance spatial reasoning and human-AI collaboration in urban planning~\cite{li2024urbangpt,feng2024citygpt}. Although LLMs have not yet been extensively adopted in GRL, their proven efficacy in other disciplines (e.g., Embedded AI~\cite{palme} and Mathematical reasoning~\cite{ahn-etal-2024-large}) highlights critical potential that merit focused exploration 
and prospective analysis.

Despite these developments, GRL continues to face several fundamental challenges. Geospatial data emerge from highly heterogeneous modalities that vary widely in semantic richness, spatial scale, and sampling density. As a result, constructing a truly unified representation space remains difficult. Robust generalization across regions is also hindered by spatial heterogeneity, spatial scale mismatch and geographic domain shift~\cite{goodchild2021replication}, where differences in urban morphology, mobility patterns, and socioeconomic conditions introduce significant distributional discrepancies. Reliable cross-modal alignment and spatial grounding are similarly challenging, particularly in the presence of conflicting signals or missing information. Furthermore, geographic knowledge, such as topological constraints, spatial hierarchies, and physical laws, is rarely incorporated in a principled manner. The introduction of LLM-based systems adds further difficulties, including geographic hallucination, spatial biases, limited grounding, and the absence of geospatial inductive mechanisms~\cite{janowicz2025geofm}. These issues collectively highlight the need for a unified, fine-grained, and forward-looking synthesis of GRL methodologies.

\begin{table}[!b]
\centering
\caption{Comparison between our survey and related surveys.}
\setlength\tabcolsep{3 pt}
\resizebox{0.98 \linewidth}{!}{
\begin{tabular}{cccccc}
\toprule
    Survey &  Taxonomy & Data Coverage  & Methodology \\
    \midrule
\citet{Chen2024SelfsupervisedLF} & Data Type & Location,Region & Unsupervised Learning \\
   \citet{mai2022review} & Pipeline & Location & Supervised Learning \\
    \citet{jin2023spatio}& Pipeline & Spatio-temporal Data& Graph Neural Network\\
    \citet{li2022geoai} & Pipeline & Visual and Mapping Data& Supervised + Unsupervised Learning\\
    \citet{wang2020deep}& Pipeline & Spatio-temporal Data& Supervised Learning\\
    
    \midrule
    \midrule
    Ours  & Pipeline & Location,Region  & Supervised + Unsupervised Learning \\
    \bottomrule
    \end{tabular}
    }
\label{tab:survey_compare}
\end{table}

\textbf{Motivations \& Related Surveys.}
Although deep learning has been widely adopted in geospatial analysis, existing surveys provide only a fragmented view of GRL. As summarized in Table~\ref{tab:survey_compare}, previous works often focus on specific data types or methodological families. For instance, \citet{Chen2024SelfsupervisedLF} systematically reviews self-supervised learning in GeoAI across location- and region-level data, but limited to geometric primitives (points, lines, polygons), omitting multimodal integration (e.g., visual / textual data) and lacking systematic evaluation of downstream applications.
In contrast, \citet{mai2022review} discusses location encoding yet leave region-level representations unexplored.
The survey by \citet{jin2023spatio} centers on graph neural networks for urban computing, while \citet{li2022geoai} focuses primarily on visual and mapping data.
The broader review in \citet{wang2020deep} provides a useful overview of spatio-temporal learning, but its perspective predates recent advances in Transformers~\cite{transformer}, multimodal pretraining~\cite{clip}, and LLMs~\cite{deepseekai2024deepseekv3technicalreport}.
Consequently, a comprehensive and up-to-date examination that synthesizes recent developments and unifies GRL under a coherent taxonomy remains lacking.

\textbf{Our Contributions.}
The present survey addresses this gap by examining the full pipeline of geospatial representation learning, as illustrated in Figure~\ref{fig:intro}, and by proposing a fine-grained taxonomy that organizes contemporary techniques in GRL. The pipeline is structured along three core dimensions: data, methodology, and applications. This framework incorporates two distinct but complementary geospatial data paradigms: (1) location-level representations that capture point-based spatial entities and their local contextual information, and (2) region-level representations that model areal units with aggregated spatial characteristics. The three-dimensional framework systematically addresses: (i) \textit{the data perspective}, focusing on geospatial data acquisition, preprocessing, and feature engineering; (ii) \textit{the methodological perspective}, encompassing model architecture design and representation learning techniques; and (iii) \textit{the application perspective}, demonstrating the deployment of learned representations in various geospatial analysis tasks and decision support systems.

Our contributions can be summarized in three parts. 
\begin{itemize}[leftmargin=*,topsep=0em,itemsep=0em]
    \item \textit{A unified and fine-grained taxonomy for GRL}. We introduce a three-level taxonomy that integrates data modalities, representation methodologies, and fusion architectures (single-view, dual-view, multi-view), providing a coherent perspective that connects both location-level and region-level paradigms.

    \vspace{-0.2em}
    \item \textit{A synthesis of methods across two technological eras.} We trace the evolution of GRL from early deep learning approaches to recent multimodal and LLM-driven methods, emphasizing shifts in modeling strategies such as contrastive learning, graph-based representations, cross-view alignment, retrieval augmentation, and LLM-empowered reasoning.

    \vspace{-0.2em}
    \item \textit{An analytical review of geospatial data modalities and their modeling implications.} We examine the roles of key modalities (including satellite imagery, trajectories, POIs, texts, and social sensing data) and discuss how their characteristics shape GRL challenges.

    \item \textit{An outlook on GRL in the LLM era.} We assess current LLM-based geospatial systems, identify major limitations such as geographic hallucination and weak grounding, highlight emerging benchmarks, and outline promising research directions involving geospatial foundation models, mixture-of-experts architectures, and agent-based urban reasoning.
\end{itemize}


\vspace{0.5em}
\noindent \textbf{Paper Organization.}
The rest of this paper is structured as follows. Section~\ref{sec:preliminary} introduces the definitions of various fundamental data formats and provides an overview of geospatial representation learning. Section~\ref{section:data modalities} elaborates on a range of specific data modalities and their respective applications within the context of this work. In Section~\ref{sec:methodology}, a taxonomy of deep learning methodologies for geospatial representation learning is proposed, categorized into three perspectives: single-view, dual-view, and multi-view approaches. Section~\ref{sec:application} consolidates a diverse array of application scenarios, while Section~\ref{sec:challenge} highlights promising research directions and unresolved challenges for future exploration in the LLM era. Finally, Section~\ref{sec:conclusion} concludes this survey with a summary of key insights.
\section{Preliminaries}
\label{sec:preliminary}
\begin{figure}[!t] 
    \centering
    \includegraphics[width=0.99\linewidth]{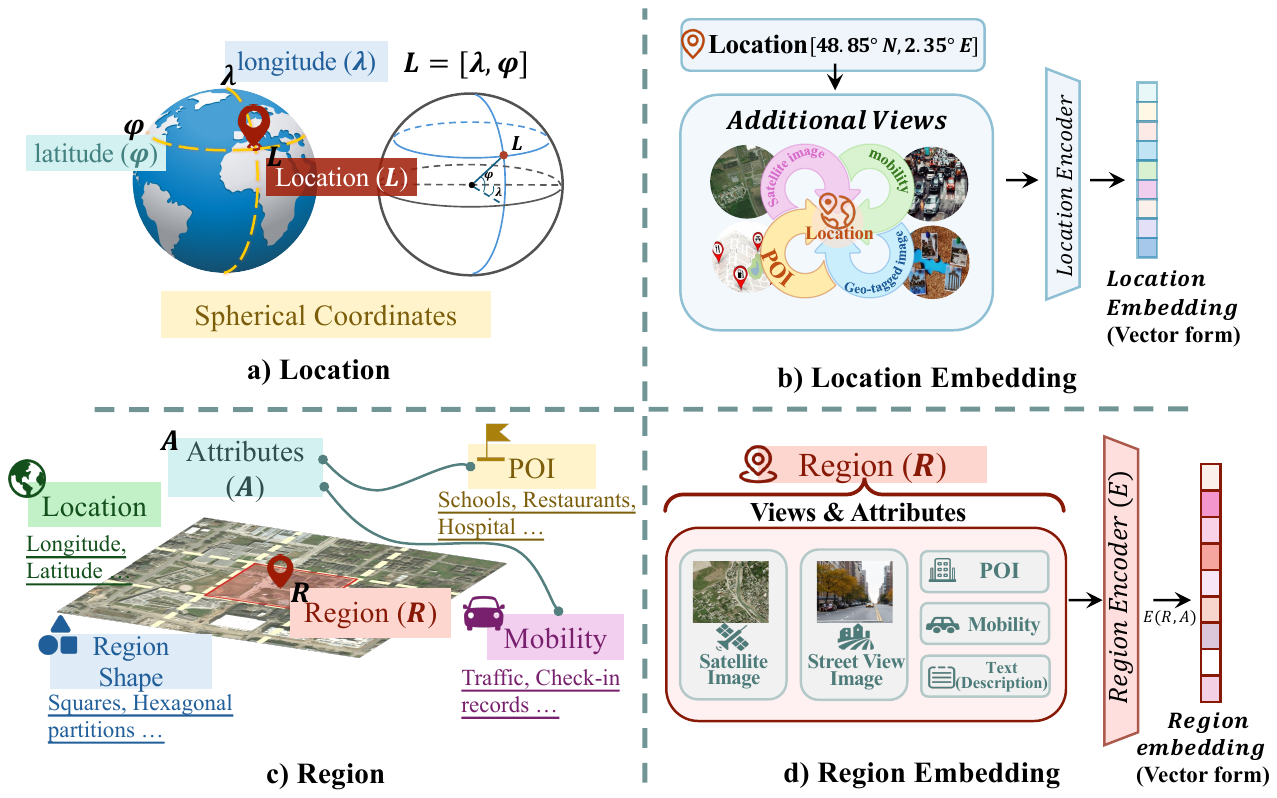}
    \caption{An illustration of concepts including location, location embedding, region, and region embedding in geospatial representation learning.}
    \label{fig:location_region}
\end{figure}

In this section, we introduce the basic formulation of this survey and provide an intuitive illustration in Figure \ref{fig:location_region}.

\subsection{Region Representation Learning} 
\noindent
\textit{Definition 1 (Urban Region).}
A city can be partitioned into a set of urban regions $\mathcal{R} = \{r_1, r_2, \dots, r_i, \dots,r_N\}$ \citep{2023KnowCL}, where $N$ denotes the number of regions in the specific city, following various criteria such as road network layouts \citep{readhan2020lightweight}, administrative boundaries \citep{urban2vec}, and sub-divisions based on specific sizes and shapes (e.g., rectangular/hexagonal grids) \citep{wang2018learning,du2019beyond,yong2024musecl}.

\vspace{0.2em} \noindent
\textit{Definition 2 (Urban Region Attribute).}
Urban Region attributes are inherent social and geographic characteristics of urban areas \citep{zhang2021multi}, which can be learned from multiple data modalities, such as POI, mobility, and urban layout (Sec.~\ref{section:data modalities}). An urban region can be comprehensively represented by a series of attributes with dimension $d$, denoted by $\mathcal{A}_i = \{a_{i,1}, a_{i, 2}, \dots, a_{i, j}, \dots, a_{i, n}\}$, where $i$ denotes the $i$-th sub-region, $j$ represents the $j$-th attribute, and $n$ denotes the number of attributes for the region.  

\vspace{0.2em} \noindent
\textit{Problem Statement 1 (Urban Region Embedding).}
The objective of urban region representation learning is to generate region embeddings with high generalizability by integrating regions and their attributes using various methodologies, such as Contrastive Learning and GNNs \citep{zou2025deep}. The process can be expressed as $v_i = \boldsymbol{M}_{R}~(\mathcal{A})$, where $\boldsymbol{M}_R$ refers to the corresponding networks. The distributed embedding $v_i$ from each sub-region $r_i$ collectively form the set of region embeddings with $D$ dimension, which can be obtained as $\mathcal{V} = \{v_1, v_2, \dots, v_N\}$, where  $\space v_i \in \mathbb{R}^{D}$.

\subsection{Location Representation Learning}

\noindent
\textit{Definition 4 (Location).}
A location can be defined as a specific position or area in space that is identified by its geographical coordinates, denoted as $\mathcal{L}=\{l_1, l_2,...,l_n\}, l_i=[\lambda_i, \phi_i]$, where $\lambda_i \in [-\pi, \pi]$ represents the longitude and $\phi_i \in [-\frac{\pi}{2}, \frac{\pi}{2}]$ represents the latitude, $n$ indicates the number of locations. 

\vspace{0.2em}
\noindent
\textit{Problem Statement 2 (Location Embedding).}
Location representation learning process is divided into two parts: coordinate vectorization and representation fusion. In coordinate vectorization, a coordinate encoder $\boldsymbol{E}~(\mathcal{L}): \mathbb{R}^{n \times 2} \rightarrow \mathbb{R}^{n \times d}$ aims to project discrete coordinates into $d$ dimensional vectors \citep{yin2021learning,mai2020multi_space2vec}. 
Since the semantic information of a geographic location does not originate from its numerical coordinate alone~\cite{satclehao2025nature}, a semantically meaningful location embedding necessitates the integration of extrinsic semantics from other data modalities.
In representation fusion, a model $\boldsymbol{M}_\mathcal{L}$  is constructed to integrate coordinate vectors with various data modalities (e.g., visual images and texts) \citep{vivanco2024geoclip, mai2023csp, satclipklemmer2023,LLMGeoveche2024geolocation} for capturing geographical and functional attributes of locations across the globe.

\section{Data Modality Perspective}
\label{section:data modalities}
This section provides a concise overview of the diverse data modalities employed in geospatial representation across various locations and regions. These data types are categorized into four primary classes: \textbf{Spatial Data}, \textbf{Mobility Data}, \textbf{Social Media Data} and \textbf{Socio-Economic Attributes}. Each category serves distinct purposes and contributes unique insights to the comprehensive understanding of geospatial dynamics.

\begin{figure}[!t] 
\centering
\includegraphics[width=0.99\linewidth]{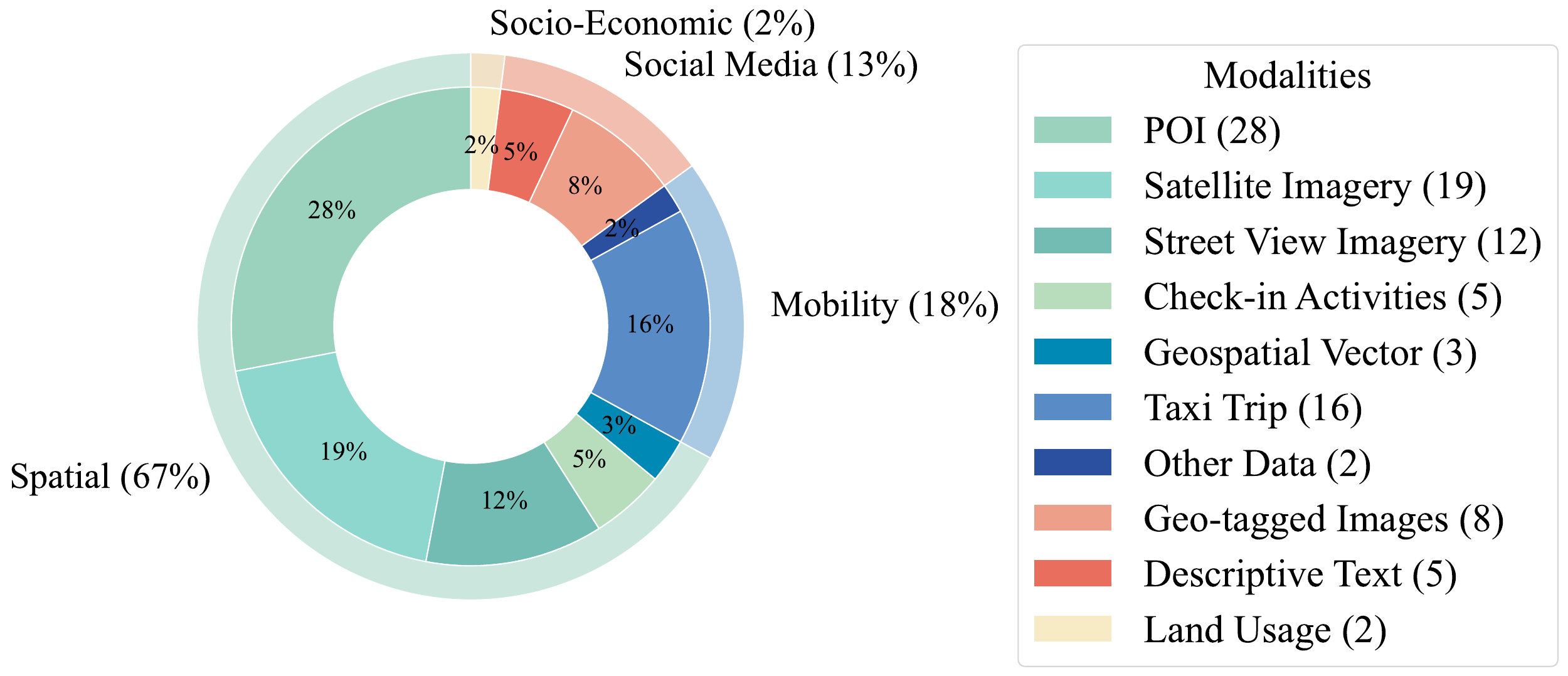}
\caption{The usage frequency of data modalities during learning stage across four categories in the survey. Each category contains commonly used data modalities.}
\label{fig:modality}
\end{figure}

\begin{figure}[!t]
\centering
\includegraphics[width=0.99\linewidth]{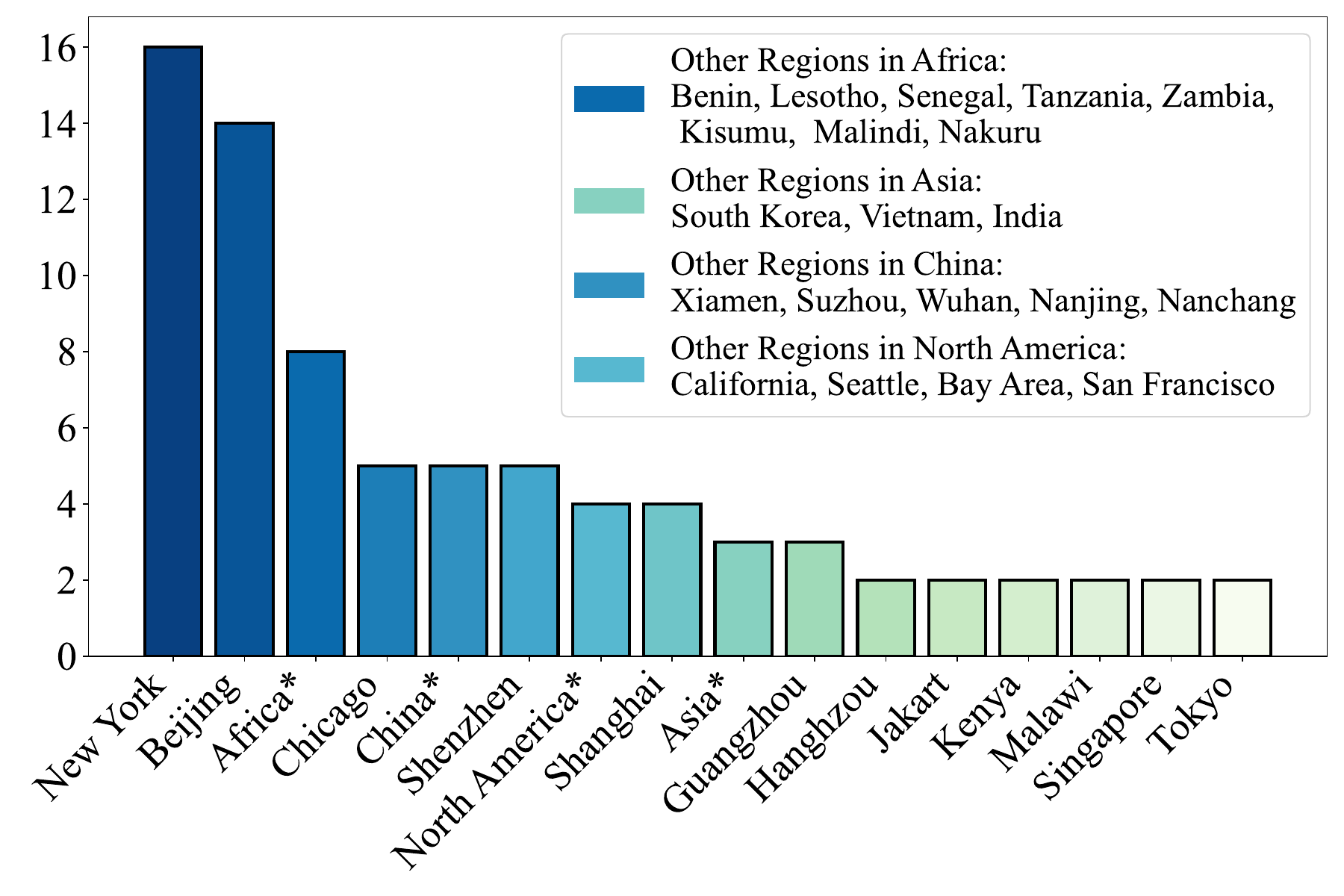}
\caption{The dataset usage frequency across cities / countries in relevant papers. Popular cities are listed individually, while other cities within a country are aggregated and marked with * denoting ``other regions'' in annotations.}
\label{fig:coverage}
\end{figure}

In Figure \ref{fig:modality}, we demonstrate the frequency of the above data categories, highlighting three primary findings. First, spatial data dominates (67\%), offering comprehensive insights via POIs and satellite imagery. Second, Mobility data ranks second (18\%), with taxi trip patterns frequently representing movement. Third, Social media data, though comprising only 13\%, holds substantial future potential for textual analysis, particularly with advancements in LLMs. Figure \ref{fig:coverage} further examines data usage in cities such as Beijing, New York, and Chicago, which leverage diverse datasets, including POIs, mobility patterns, and imagery. Developed cities receive more focus due to their well-established infrastructure and open data initiatives, while studies on African cities emphasize socio-economic metrics.

\subsection{Spatial Data}
\noindent
Spatial data identifies the geographic location and characteristics of objects on the Earth's surface, which can be used for describing the details of location and region, including:  
\begin{itemize}[leftmargin=*,itemsep=0em]
    \item \textbf{Points of Interest (POIs)} represent a collection of specific locations or sites of significance \citep{Wikipedia_POI}. It can be denoted as $\mathcal{P}^{r} = \{p_1, p_2, \dots, p_m\}$, where $\mathcal{P}^{r}$ represents a set of POIs, and $m$ denotes the number of POIs within a region $r$. Formally, each POI, $p_i = [n_i, l_i, c_i, a_i]$, contains the name $n_i$, coordinates $l_i$, category $c_i$ and additional attributes $a_i$, where the category is selected from a hierarchical taxonomy that includes major categories and corresponding subcategories.
    
    \item \textbf{Check-in Activities} as a region attribute \citep{zhang2021multi}, are generated by users at specific POIs \citep{MVURE}, incorporating human activities and reflecting urban dynamics. 
    Each check-in record can be formally represented as a triplet $(u,p,t)$, where $u$ denotes the user, $p$ represents the POI, and $t$ indicates the timestamp.

    \item \textbf{Satellite Imagery} provides a bird's-eye view of Earth's surface, capturing urban layouts, environments and building distribution \citep{yong2024musecl}. Each city or region can be segmented into multiple satellite-image tiles, denoted by $\mathcal{SI}= \{I^{sa}_1, I^{sa}_2, \dots, I^{sa}_n\},~I^{sa}_i \in \mathbb{R}^{H \times W \times C}$, where $n$ denotes the total number of satellite images, and $H$, $W$, and $C$ represent the height, width, and number of channels of each imagery, respectively.

    \item \textbf{Street View Imagery} is a photograph that can be captured from ground-level perspectives along streets and roads, providing comprehensive visual clues at street level, denoted by $\mathcal{SV} = \{I^{sv}_{1}, I^{sv}_{2}, \dots, I^{sv}_{m}\}, ~I^{sv}_i \in \mathbb{R}^{H \times W \times C} $, where $m$ denotes the number of street view imagery. The definitions of $H$, $W$, and $C$ are the same as those in satellite imagery.

    \item \textbf{Geospatial Vector} represents geographic features such as points, polylines, and polygons, capturing spatial relationships and attributes. In OpenStreetMap (OSM), these vectors are defined by  nodes (i.e., $p_i$), ways (i.e., $w_i = [p_1, p_2, \dots, p_m]$), relations (i.e., $e_i = [w_1, w_2, \dots, w_n]$), and building footprints \citep{hafusionsun2024urban, cityfmbalsebre2024}. 
\end{itemize}

\subsection{Mobility Data}

Mobility data reflects humans' transitions behavior among POIs within cities \citep{managing, mvpnfu2019efficient}, recorded as a sequence of points with geographic coordinates and timestamps, which is often represented by taxi trip \citep{du2019beyond, mvpnfu2019efficient, cgalzhang2019unifying}. Taxi trip is generated within urban areas, typically related to urban dynamics, denoted by a set of trajectories $\mathcal{T} = \{\tau_1, \tau_2, \dots, \tau_n\}$. Each trip $\tau_i$ contains $[p_s, p_e; t_s, t_e]$, where $p_s$ and $p_e$ represent the starting and ending geographic coordinates (i.e., latitude and longitude), and $t_s$ and $t_e$ display the starting and ending times of the trip, respectively \citep{hdgewang2017region}. Beyond taxi data, other sources such as subway ridership \citep{SEONG2023251} are also leveraged to represent mobility.

\subsection{Social Media Data}


The proliferation of social media and location-based platforms (e.g., Twitter / X, Facebook) has fostered the crowdsourcing of extensive geospatial data through user-generated, geo-tagged content \citep{zou2025deep}. This multimodal social sensing data has fueled interdisciplinary advances in multimodal representation learning~\citep{cheng2022vista, chen2024internvl}.

\begin{itemize}[leftmargin=*,itemsep=0em]
    \item \textbf{Geo-tagged Imagery} refers to images linked with metadata  from platforms like social media (e.g., YouTube, Facebook), open mapping services (e.g., Google Maps), which contains geographic coordinates, timestamps, identifiers, etc., \citep{yin2021gps2vec, yin2021learning, mai2022sphere2vec, mai2023csp}. 

    \item \textbf{Descriptive Texts} sourced from online encyclopedias (e.g., websites \citep{urban2vec}, Wikipedia \citep{uzkent2019learning}), and generative models \citep{urbanclip, hao2024urbanvlp}, provide contextual details about locations and entities~\citep{sheehan2019predicting}. 
    Furthermore, user-generated reviews provide granular insights into human perception and urban vibrancy. 
    They could capture collective sentiment and functional interactions, serving as essential real-time proxies for monitoring socioeconomic vitality and understanding the human-centric dimensions of geospatial spaces.
    Besides, the emergence of large language models (LLMs)\citep{openai2024gpt4technicalreport, deepseekai2024deepseekv3technicalreport} further enriches textual descriptions, advancing multimodal alignment and representation learning~\citep{zou2025deep, li2024urbangpt}.

\end{itemize} 
\clearpage

\subsection{Socio-Economic Attributes}

Socio-economic attributes quantify the interplay between human populations, economic activities, and their physical environments. Unlike raw spatial or mobility data, these attributes often serve as high-level semantic indicators used to evaluate the effectiveness of learned representations in downstream tasks~\cite{WANG2022102572}.

Common socio-economic attributes employed in geospatial representation tasks include \textit{land usage} \citep{steurer2020measuring,risal2020sensitivity,chen2021mapping}, \textit{demographic data} \citep{learningscore}, \textit{crime statistics} \citep{huang2018deepcrime}, \textit{income} \citep{hdgewang2017region}, \textit{poverty} \citep{jean2019tile2vec}, \textit{check-in activities} \citep{mgfn}, \textit{nightlight imagery} \citep{keola2015monitoring}, \textit{house price} \citep{hao2024urbanvlp}, and \textit{carbon emissions} \citep{urbanclip}. They often serve as predictive indicators in downstream tasks to assess the performance of geospatial representations. 
Other data such as the \textit{number of takeaways (and reviews)} \citep{pgsimclrxi2022beyond}, \textit{women's BMI} \citep{sceneparselee2021predicting}, and \textit{Origin-Destination Employment Statistics (OSRM)} \citep{liu2020learning} also offer unique insights into urban development.

\section{Methodology Perspective}
\label{sec:methodology}
\subsection{Data-centric View}

\definecolor{mycolor}{RGB}{215, 245, 200}

\tikzstyle{leaf}=[
    draw=hiddendraw,
    rounded corners,
    minimum height=1em,
    fill=mycolor,
    text opacity=1, 
    align=center,
    fill opacity=.5,  
    text=black,
    align=left,
    font=\footnotesize,
    inner xsep=3pt,
    inner ysep=1pt,
]

\tikzstyle{middle}=[
    draw=hiddendraw,
    rounded corners,
    minimum height=1em,
    fill=output-white!40,
    text opacity=1, 
    align=center,
    fill opacity=.5,  
    text=black,
    align=left,
    font=\footnotesize,
    inner xsep=3pt,
    inner ysep=1pt,
]

\begin{figure*}[t!]
    \vspace{-1em}
    \centering
    \begin{forest}
        for tree={
            forked edges,
            grow=east,
            reversed=true,
            anchor=base west,
            parent anchor=east,
            child anchor=west,
            base=middle,
            font=\footnotesize,
            rectangle,
            line width=0.7pt,
            draw=output-black,
            rounded corners,
            align=left,
            minimum width=2em,
            s sep=5pt,
            inner xsep=3pt,
            inner ysep=1pt,
        },
        where level=1{text width=4.5em}{},
        where level=2{text width=6em,font=\footnotesize}{},
        where level=3{font=\footnotesize}{},
        where level=4{font=\footnotesize}{},
        where level=5{font=\footnotesize}{},
        [Geospatial Representation Learning, middle, rotate=90, font=\normalsize, anchor=north, edge=output-black
            [Single View, middle, edge=output-black, text width=5.3em
                [Single View \\Location Embedding, middle, text width=7em, edge=output-black
                    [Loc2Vec~\citep{spruyt2018loc2vec}{,} Place2Vec~\citep{place2vecyan2017itdl}{,} Translocator~\cite{translocator}, leaf, text width=17.5em, edge=output-black]
                ]
                [Single View \\Region Embedding, middle, text width=7em, edge=output-black
                    [Satellite \\Imagery
                        [Tile2Vec~\citep{jean2019tile2vec}{,} READ~\citep{readhan2020lightweight}{,} \citep{han2020learning}{,} \citep{yeh2020using}{,} \citep{li2023point}, leaf, text width=15em, edge=output-black]
                    ]
                    [Mobility
                        [HDGE~\citep{hdgewang2017region}{,} ZE-Mob~\citep{zemobyao2018representing}{,} MGFN~\citep{mgfn}, leaf, text width=14.8em, edge=output-black]
                    ]
                    [Others
                        [SceneParse~\citep{sceneparselee2021predicting}{,} HGI~\citep{hgihuang2023learning}{,} MTE~\citep{mtezhang2024urban}{,} CityFM~\citep{cityfmbalsebre2024}, leaf, text width=17.5em, edge=output-black]
                    ]
                ]
            ]
            [Dual View, middle, edge=output-black, text width=5.3em
                [Dual View \\Location Embedding, middle, text width=7em, edge=output-black
                    [Location \\+ Others
                        [Visual Imagery
                            [Satellite Imagery, middle, edge=output-black
                                [Geo-SSL~\citep{geosslayush2021geography}{,} CSP~\citep{mai2023csp}{,} SatCLIP~\citep{satclipklemmer2023}{,}\\ 
                                 SatCLE~\citep{satclehao2025nature}{,} TorchSpatial~\citep{wu2024torchspatial}{,}\\ 
                                 RANGE~\citep{dhakal2025range}{,} \cite{pan2025largescalegeolocalization}, leaf, text width=13em, edge=output-black]
                            ]
                            [Geo-Tagged Imagery, middle, edge=output-black
                                [GPS2Vec~\citep{yin2021gps2vec}{,} GPS2Vec+~\citep{yin2021learning}{,} \\ 
                                 Sphere2Vec~\citep{mai2022sphere2vec}{,} GeoCLIP~\citep{vivanco2024geoclip}, leaf, text width=11.7em, edge=output-black]
                            ]
                        ]
                        [Text, middle, edge=output-black
                            [GeoLLM~\citep{manvi2023geollm}{,} LLMGeovec~\citep{LLMGeoveche2024geolocation}, leaf, text width=11em, edge=output-black]
                        ]
                        [Trajectory
                            [TALE~\citep{talewan2021pre}{,} \citep{park2023pre}, leaf, text width=7.2em, edge=output-black]
                        ]
                        [Geographic Context
                            [MGeo~\citep{ding2023mgeo}, leaf, text width=4em, edge=output-black]
                        ]
                    ]
                ]
                [Dual View \\Region Embedding, middle, text width=7em, edge=output-black
                    [POI + Mobility
                        [\citep{du2019beyond}{,} MV-PN~\citep{mvpnfu2019efficient}{,} CGAL~\citep{cgalzhang2019unifying}{,} Region2Vec~\citep{luo2022region2vec}{,} MVURE~\citep{MVURE}{,} \\
                         ROMER~\citep{romer}{,} HREP~\citep{hrep}{,} EUPAC~\citep{chan2024enhanced}{,} RECP~\citep{recpli2024urban}{,} ReMVC~\citep{zhang2022region}{,} \\
                         URGent~\citep{urgenthou2022urban}{,}~\citep{zhang2025urban}, leaf, text width=23.5em, edge=output-black]
                    ]
                    [Visual Imagery \\+ Others
                        [Satellite Imagery
                            [POI
                                [PG-SimCLR~\citep{pgsimclrxi2022beyond}{,} MMGR~\citep{mmgrbai2023geographic}{,} \\ 
                                 ReFound~\citep{xiao2024refound}{,} GeoHG~\citep{geohgzou2024learning}, leaf, text width=13em, edge=output-black]
                            ]
                            [Knowledge \\Graph
                                [KnowCL~\citep{2023KnowCL}, leaf, text width=5em, edge=output-black]
                            ]
                            [Text
                                [UrbanCLIP~\citep{urbanclip}, leaf, text width=7.3em, edge=output-black]
                            ]
                        ]
                        [Street-view Imagery
                            [POI
                                [Urban2Vec~\citep{urban2vec}, leaf, text width=6.25em, edge=output-black]
                            ]
                            [Text
                                [USPM~\citep{uspmchen2024profiling}, leaf, text width=4em, edge=output-black]
                            ]
                        ]
                        [Satellite Imagery \\+ Street-view Imagery
                            [\citep{li2022predicting}, leaf, text width=1.5em, edge=output-black]
                        ]
                    ]
                ]
            ]
            [Multiple View, middle, edge=output-black, text width=5.3em
                [Multiple View \\Location Embedding, middle, text width=7em, edge=output-black
                    [UrbanFusion~\citep{muhlematter2025urbanfusion}{,} AETHER~\citep{aether}{,} GT-Loc~\citep{shatwell2025gtloc}{,} GAIR~\citep{liu2025gair}{,} PIGEON~\cite{haas2024pigeon}{,}\\ 
                     \cite{lindenberger2025scalingimagegeolocalization}{,} \cite{jia2025towardsinterpretable}{,} GeoDecoder~\cite{geodecoder}{,} CityGuessr~\cite{kulkarni2024cityguessr}{,} G3~\cite{jia2024g3}{,} Georanker~\cite{georanker}{,} \\
                     GLOBE~\cite{globe}, leaf, text width=26em, edge=output-black]
                ]
                [Multiple View \\Region Embedding, middle, text width=7em, edge=output-black
                    [RegionEncoder~\citep{regionencoderjenkins2019unsupervised}{,} M3G~\citep{m3ghuang2021learning}{,} Geo-Tile2Vec~\citep{geotile2vecluo2023geo}{,} MuseCL~\citep{yong2024musecl}{,}\\ 
                     UrbanVLP~\citep{hao2024urbanvlp}{,} HAFusion~\citep{hafusionsun2024urban}{,} HUGAT~\citep{kim2022hugat}{,} HealthCLIP~\citep{healthclipouyang2024}{,}\\
                     FlexiReg~\citep{sun2025flexireg}{,} GURPP~\citep{gurpp}{,} ~\cite{namgung2025lessismore}{,} MobCLIP~\citep{wen2025mobclip}{,} GraphJCL~\citep{zhao2025graphjcl}{,} ~\citep{zhao2025modality}{,}\\
                     UrbanLN~\cite{urbanln}{,} UrbanMMCL~\cite{UrbanMMCL}{,} MGRL4RE~\cite{MGRL4RE}, leaf, text width=26em, edge=output-black]
                ]
            ]
        ]
    \end{forest}
    \caption{A taxonomy of representative works for geospatial representation learning.}
    \label{fig:taxonomy}
\end{figure*}
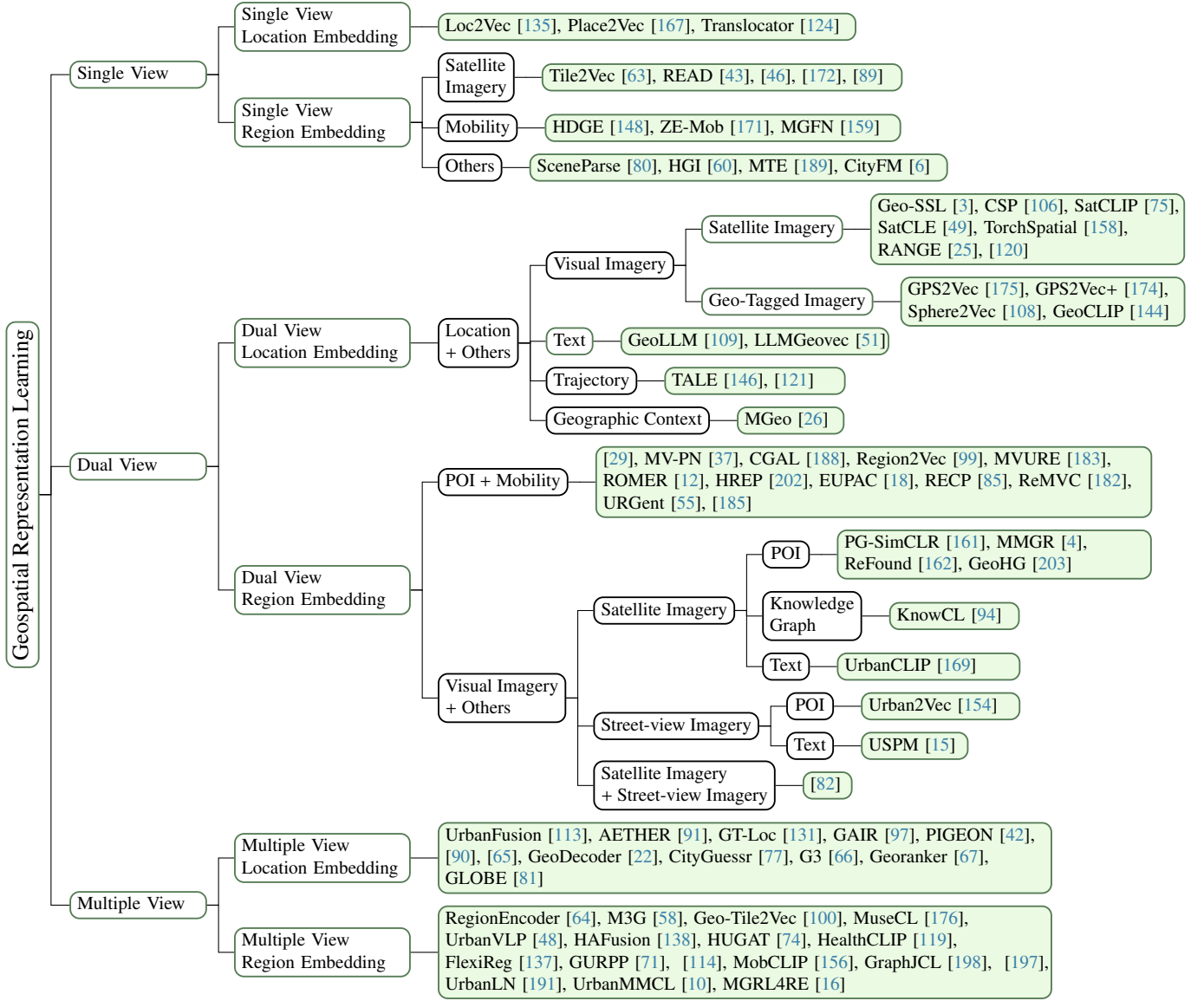


Cross-modal data integration~\citep{zheng2014urban,zou2025deep,ufmsurvey-kdd2024,Chen2024SelfsupervisedLF} is pivotal for shaping global or city-level geospatial embeddings. We systematically review recent advancements in geospatial representation learning (i.e., geospatial embedding), analyzing their evolution and classifying methods into \textbf{single-view}, \textbf{dual-view}, and \textbf{multiple-view} approaches based on the complexity of views in learning geospatial representations.
A taxonomy of representative work is provided in Figure~\ref{fig:taxonomy}, with detailed summaries in Table~\ref{tab:summarizations}. 
We organize our discussion from location-level and region-level views respectively.
\subsubsection{Single View}



In this section, we systematically review \textit{Single View Geospatial Embedding} methods, which project diverse spatial data modalities into a unified representational space.




\vspace{0.2em}
\textbf{\textit{Single View Location Embedding.}}
Single view location embedding refers to the process of extracting and representing information centered around specific \textbf{geographical coordinates}.
Loc2Vec~\citep{spruyt2018loc2vec} represents one of the earliest efforts to capture location semantics using environmental contexts retrieved from GIS queries.
Place2Vec~\citep{yan2017itdl} uses the distributional differences of POI types as semantic information to enhance place embeddings.

\vspace{0.2em}
\textbf{\textit{Single View Region Embedding.}}
Similarly, single-view region embeddings extract and refine the single-modal attributes associated with their corresponding urban regions.
\begin{itemize}[leftmargin=*,itemsep=0.1em, parsep=0.2em, topsep=0.2em]
    \item \textbf{1) Satellite Imagery}: Satellite visual data extraction is the most significant component in single-view region embedding due to its global coverage, accessibility~\citep{zou2025deep}, and ability to capture diverse physical, environmental, and socio-economic features. READ~\citep{readhan2020lightweight} pioneers a semi-supervised approach using the mean-teacher model~\citep{tarvainen2017mean} to analyze satellite imagery, with a specific emphasis on economic scales in South Korea. \cite{han2020learning} further integrates the partial order graph to cluster satellite imagery, thus facilitating the assessment of economy.

    \vspace{0.2em}
    \item \textbf{2) Mobility}. Mobility flow~\citep{hdgewang2017region,zemobyao2018representing,mgfn} captures geospatial semantics dynamically through movement patterns, revealing spatial interactions, temporal variations, and socio-economic traits.
    HDGE~\citep{hdgewang2017region} employs a heterogeneous graph structure to simultaneously account for temporal dynamics and multi-hop location transitions. 
    MGFN~\citep{mgfn} prioritizes temporal information by constructing mobility graphs per timestep, aggregating similar graphs to extract varied mobility patterns.
    \item \textbf{3) Other Modalities.}
    Other modalities~\citep{sceneparselee2021predicting,hgihuang2023learning,mtezhang2024urban,cityfmbalsebre2024} that capture spatial contexts and dynamic characteristics have also been explored and utilized for modeling socio-economic attributes. However, these approaches have not yet gained widespread adoption.
HGI~\citep{hgihuang2023learning} exclusively leverages POI data to model regional features, employing GCNs to hierarchically aggregate POI embeddings from the region to city level.
MTE~\citep{mtezhang2024urban} represents trajectories via transition, spatial, and temporal views, effectively capturing socio-economic characteristics and aiding land use type prediction.

\end{itemize}

\subsubsection{Dual View}

\begin{figure}[!t]
    \centering
    \includegraphics[width=0.99\linewidth]
    {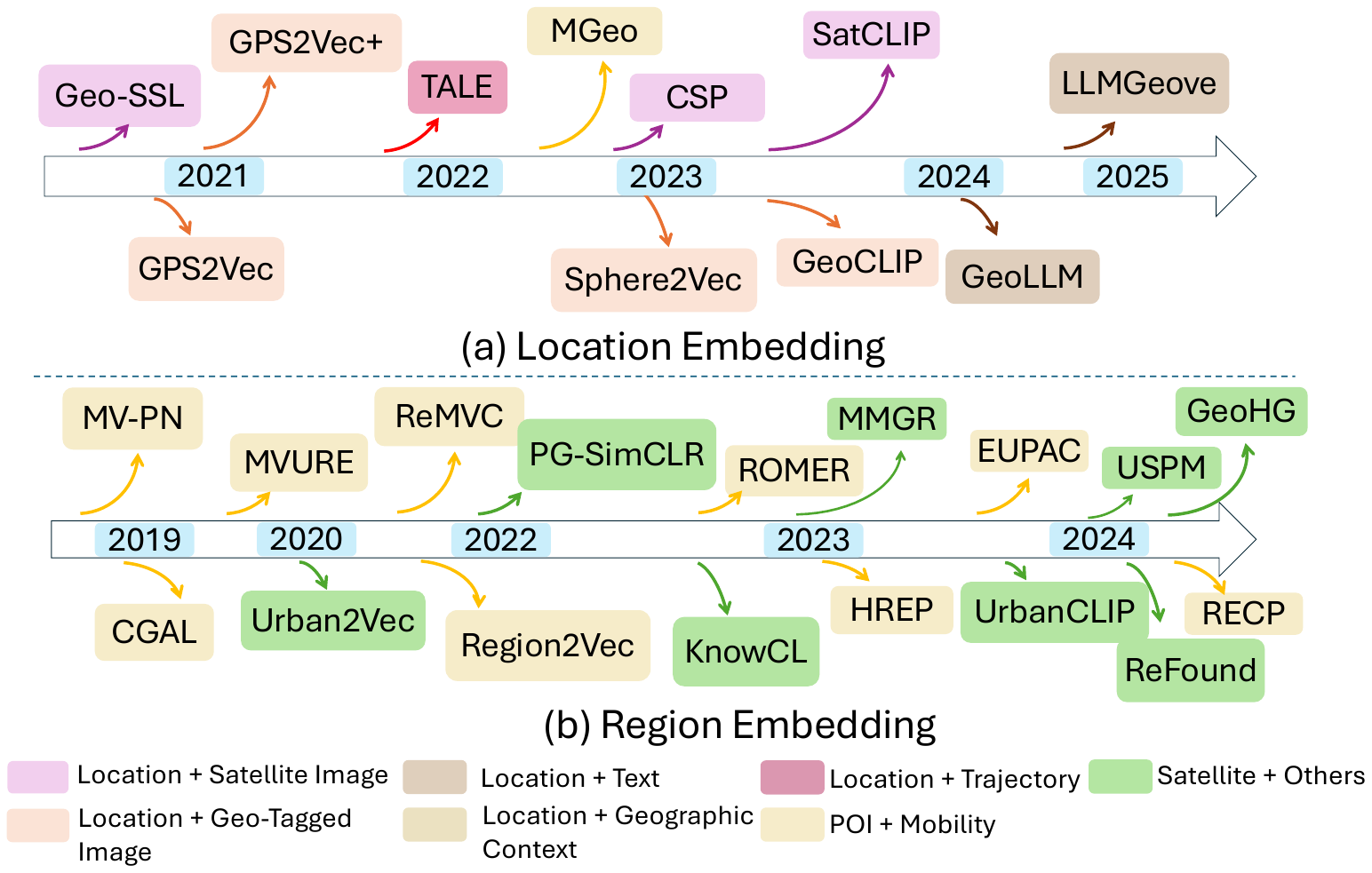} 
     \caption{The roadmap of Dual View.}
    \label{fig:roadmap}
\end{figure}

The dual view perspective integrates data from two distinct modalities to address the limitations of single-modality frameworks. For example, visual data provides detailed spatial representations, whereas mobility flow captures dynamic processes and temporal variations. Figure~\ref{fig:roadmap} illustrates the temporal development trends of dual
view approaches.

\vspace{0.2em}
\textbf{\textit{Dual View Location Embedding.}}
In summary, the dual view location embedding presents a \textbf{location + others} pattern, which integrates geographic location data with various multidimensional modalities, facilitating a deeper understanding of the complex interactions. 
In this pattern, \textit{visual data} continues to play a predominant role. By integrating geographic location information with satellite imagery~\citep{geosslayush2021geography,mai2023csp,satclipklemmer2023} and geo-tagged imagery~\citep{yin2021gps2vec,yin2021learning,mai2022sphere2vec,vivanco2024geoclip}, we can obtain a comprehensive representation of land use types, changes in natural resources and socio-economic conditions.

In addition to visual data, the representation of \textit{other forms of information} also merits considerable attention. MGeo~\citep{ding2023mgeo}, designed for query-POI matching, treats geographic context as an independent modality, emphasizing spatial relationships 
between a point and its surroundings.
As large language models (LLMs) gain widespread attention~\citep{zhao2023survey,kumar2024large},
GeoLLM~\citep{manvi2023geollm} and LLMGeovec~\citep{LLMGeoveche2024geolocation} both integrate nearby location information from OSM to construct textual prompts, enabling the extraction of geospatial knowledge.


\vspace{0.2em}
\textbf{\textit{Dual View Region Embedding.}}
From the perspective of explicit region characterization, the dual view exemplifies two distinctive developmental trajectories of model architectures: \textbf{POI + Mobility}~\citep{du2019beyond,mvpnfu2019efficient,cgalzhang2019unifying,luo2022region2vec,wang2018learning,C-MPGCN,SEONG2023251,MVURE,romer,hrep,chan2024enhanced,recpli2024urban,zhang2022region} and \textbf{Visual Imagery + Others}~\citep{pgsimclrxi2022beyond,mmgrbai2023geographic,xiao2024refound,2023KnowCL,yan2024urbanclip,urban2vec,uspmchen2024profiling,zhang2024multi,li2022predicting}.
Next, we discuss each of them in detail.

\begin{itemize}[leftmargin=*,itemsep=0.1em, parsep=0.2em, topsep=0.2em]
\item \textbf{1) POI + Mobility.}
In urban region embedding, POIs offer static functional attributes, while mobility data reflects dynamic activity patterns. Their integration facilitates the modeling of spatiotemporal characteristics, enhancing urban dynamic analysis.
Based on a comprehensive literature review, we summarize that the primary distinction across relevant works lies in the approaches employed for mobility graph construction. Studies such as \citep{du2019beyond,mvpnfu2019efficient,cgalzhang2019unifying,luo2022region2vec} utilize geographical distance and human mobility to construct multi-view graphs, which are subsequently integrated through techniques like autoencoders~\citep{du2019beyond,mvpnfu2019efficient} and graph convolutional networks (GCNs)~\citep{cgalzhang2019unifying,luo2022region2vec}.
MVURE~\citep{MVURE} and ROMER~\citep{romer} integrate additional graph perspectives, such as check-in and source / destination graphs, employing attention mechanisms to fuse information across multiple modules.
\cite{hrep,chan2024enhanced} explore heterogeneous graphs in region embedding, using multiple edge types to represent diverse node relationships.




Although most works are grpah-based, notable exceptions include ReCP~\citep{recpli2024urban} and ReMVC\citep{zhang2022region}, which deviate from graph-based methods by representing both dynamic and static attributes of regions using POI distributions and inflow/outflow counts, enhanced by contrastive learning.

\item \textbf{2) Visual Imagery + Others.}
Here we discuss scenarios where satellite imagery and street-level imagery are each combined with other modalities as visual data.

\textbf{Satellite Imagery.} The semantic richness of satellite imagery enables it to be enhanced by various multimodal data to precisely characterize regions. 
Compared to other modalities, \textit{\textbf{POI}} data, characterized by its fine-grained categorization, strong spatial alignment, and ease of acquisition, can be effectively integrated with satellite imagery to enhance socio-economic representation.
PG-SimCLR~\citep{pgsimclrxi2022beyond} and MMGR~\citep{mmgrbai2023geographic} are grounded in the representation of physical and geographic information derived from satellite imagery, leveraging POI categories as a quantitative proxy for human activity factors. 
ReFound~\citep{xiao2024refound} transforms POI data and satellite imagery into unified embeddings using knowledge distillation from multiple pre-trained foundation models, transferring their generalization capabilities to urban region modeling.

KnowCL~\citep{2023KnowCL} pioneers the use of \textit{\textbf{Knowledge Graphs (KGs)}} to model urban knowledge. By introducing image-KG pairs, it enhances semantic-visual representation alignment through mutual information maximization.
With the surge of LLMs across various domains~\citep{zhao2023survey,kumar2024large}, the interpretability of text has garnered significant attention. UrbanCLIP~\citep{yan2024urbanclip} leverages \textit{\textbf{LLMs}} to generate descriptions of satellite imagery. 

\textbf{Street-view Imagery.}
The advantage of street-view imagery lies in its ability to provide fine-grained semantic information at the location level.
Urban2Vec~\citep{urban2vec}  models POI data using a bag-of-words~\citep{harris1954distributional} approach and employs contrastive learning to align it with features derived from street-view imagery.
Aimed at urban street profiling, USPM~\citep{uspmchen2024profiling}  integrates street imagery with textual information and employs semi-supervised graph learning based on spatial topology.
\cite{li2022predicting} comprehensively investigates the distinct roles and complementary functions of visual modalities (e.g., satellite imagery and street-view imagery) at various levels in urban region representation learning.
\end{itemize}

\subsubsection{Multiple View}
Although the dual view paradigm currently dominates the geospatial representation learning field, the diverse and flexible data modalities in this domain still offer opportunities for enhanced representations through additional information. However, the trade-off between the cost of incorporating new modalities and the performance improvements requires evaluation.

Researchers have creatively combined the two modeling pipelines within the dual view framework. For example, RegionEncoder~\citep{regionencoderjenkins2019unsupervised} is the first to integrate satellite imagery, POIs, and human mobility data to jointly learn region representations through GCN and denoising autoencoder. To more effectively model POI intra-relationships, both M3G~\citep{m3ghuang2021learning} and Geo-Tile2Vec~\citep{geotile2vecluo2023geo} incorporates street-level visual data in conjunction with POIs and mobility information.
UrbanVLP~\citep{hao2024urbanvlp} takes a different approach by utilizing textual descriptions as a substitute for the functionality of POIs and mobility data. It incorporates the web-scale knowledge compressed within LLMs into region embeddings with effective quality control.

In addition to POIs and mobility data, \textit{\textbf{Land Usage}} also serves as an auxiliary feature that can provide significant semantic information for socio-economic tasks.
HAFusion~\citep{hafusionsun2024urban} incorporates land usage as a third perspective, in addition to POIs and mobility, to comprehensively capture region features from three distinct angles. An attentive fusion mechanism is employed to facilitate information interaction across different modalities and effectively integrate these multi-view representations.




\subsection{Representation Learning Methodology / Implementation}

\subsubsection{Location Embedding Methodology}


\begin{itemize}[leftmargin=*,itemsep=0.1em, parsep=0.2em, topsep=0.2em]
\item \textbf{1) \textit{Contrastive Learning.}}
As a classical unsupervised learning method, contrastive learning~\cite{simclr,moco} aims to learn effective feature representations by comparing similarities and differences between data samples. Its core idea is to enable the model to pull similar samples (positive pairs) closer in the feature space while pushing dissimilar samples (negative pairs) apart, thereby capturing the underlying structure of the data.
In contrastive learning, the most commonly used loss function is the InfoNCE~\cite{clip} loss, which is formulated as follows:
\begin{equation}
\setlength\abovedisplayskip{0.5em}
\setlength\belowdisplayskip{0.5em}
    L = -\log \frac{\exp(\text{sim}(z_i, z_j)/\tau)}{\sum_{k=1}^N \exp(\text{sim}(z_i, z_k)/\tau)},
\end{equation}
where ($z_i$, $z_j$) are the embeddings of a positive sample pair, ($z_i$, $z_k$) are the embeddings of a negative sample pair,  and $\tau$ is the temperature coefficient.

In addition, the triplet loss is widely utilized, which compares three samples (anchor, positive, and negative samples) to optimize the distances within the feature space. The structure of triplets allows for more flexible sample selection, 
especially in the case of data imbalance.
\begin{equation}
\setlength\abovedisplayskip{0.5em}
\setlength\belowdisplayskip{0.5em}
    L = \max\left( d(A, P) - d(A, N) + margin, 0 \right),
\end{equation}
where $A$ is the anchor point, $P$ is the positive sample, and $N$ is the negative sample.
$d (A, P)$ denotes the distance between the anchor point and the positive sample.
$d (A, N)$ denotes the distance between the anchor point and the negative sample.
$margin$ is a hyperparameter that ensures separation between positive and negative samples.

Contrastive learning, owing to its conceptual simplicity and versatile applicability, has emerged as a pivotal approach for intra-modal representation learning and inter-modal information alignment.
Particularly, driven by the rise of image-text multimodal paradigms~\cite{clip}, it has become the method of choice for integrating visual data with other modalities.
In the single-view scenario, Loc2Vec~\cite{spruyt2018loc2vec} employs a triplet loss framework to effectively encode geo-spatial relationships and semantic similarities that characterize the surroundings of a given location.
Within the dual-view context, \cite{geosslayush2021geography,mai2023csp,satclehao2025nature,satclipklemmer2023,vivanco2024geoclip} all endow digital location coordinates with semantic information through contrastive learning between locations and visual imagery.

\item
\textbf{2) \textit{Large Language Models.}}
The vast amount of world knowledge of LLMs has established them as an important approach for data fusion~\cite{zou2025deep}.
In location embedding, due to the lack of geo-foundation models and sufficient data volumes, existing work remains focused on extracting geographical information stored within LLMs, using various prompts and meta-information~\cite{manvi2023geollm,LLMGeoveche2024geolocation}.
GeoLLM~\cite{manvi2023geollm} endows the LLM with specific geo-contextual information from the vicinity of a location (obtained from OpenStreetMap) and then uses it to predict downstream indicators like population and income.
Instead of directly predicting indicators, LLMGeovec~\cite{LLMGeoveche2024geolocation} acquires intermediate embeddings from LLM, which are subsequently utilized to augment time series forecasting and spatial-temporal forecasting.

\item
\textbf{3) \textit{Others.}}
Beyond the two previously discussed embedding methods that have become prevalent, alternative representation learning paradigms exist. While the volume of research utilizing these approaches is comparatively modest, they are indispensable for guaranteeing modeling flexibility and adaptability.
RANGE~\cite{dhakal2025range}, G3~\cite{jia2024g3} and Georanker~\cite{georanker} utilize \textbf{retrieval augmented matching} to enhance geo-embedding.
UrbanFusion~\cite{muhlematter2025urbanfusion} and MGeo~\cite{ding2023mgeo} leverage \textbf{transformer encoder and masking strategy} for feature modeling and interaction.
GLOBE~\cite{globe} finetunes LVM via \textbf{reinforcement learning} to improve the performance of geo-localization.

\end{itemize}
\vspace{-0.5em}

\subsubsection{Region Embedding Methodology}

\begin{itemize}[leftmargin=*,itemsep=0.1em, parsep=0.2em, topsep=0.2em]
    \item 

\textbf{1) \textit{Contrastive Learning.}}
Similar to the \textit{Location Embedding Methodology}, Contrastive Learning within the \textit{Region Embedding Methodology} has been widely applied across works spanning single-view, dual-view, and multi-view perspectives.
In the single-view scenario, Tile2vec~\cite{jean2019tile2vec} employs triplet loss to differentiate features of neighboring and non-neighboring satellite images within a single data domain. CityFM~\cite{cityfmbalsebre2024} utilizes three types of contrastive objects from OSM data—nodes, polylines, and polygons—as well as relational information.

In the dual-view setting, PG-SimCLR~\cite{pgsimclrxi2022beyond}, MMGR~\cite{mmgrbai2023geographic}, KnowCL~\cite{2023KnowCL}, and UrbanCLIP~\cite{urbanclip} respectively perform contrastive learning on satellite imagery to align and integrate information with different modalities such as POI, Knowledge Graph, and Text, with similar model architectures.
Similarly, Urban2Vec~\cite{urban2vec} and USPM~\cite{uspmchen2024profiling} perform contrastive learning between street view images and POIs as well as text.

In the multiple view, MuseCL~\cite{yong2024musecl} and UrbanVLP~\cite{hao2024urbanvlp} jointly incorporate satellite and street-view imagery, where the former aligns heterogeneous urban data with POI data and mobility patterns through contrastive learning, while the latter establishes semantic correlations with synthesized textual descriptions via cross-modal alignment.

\item
\textbf{2) \textit{Graph Neural Networks.}}
Graph Neural Network (GNN) represents a type of deep learning model specifically designed for processing graph-structured data, capable of effectively capturing complex relationships between nodes and non-local dependencies.
The core idea of GNNs is to aggregate neighborhood information and update node representations through \textit{Message Passing}. 
Graph Convolutional Network (GCN) is one classical GNN that propagates information through a normalized adjacency matrix.
\begin{equation}
H^{(l+1)} = \sigma\left(\tilde{D}^{-\frac{1}{2}} \tilde{A} \tilde{D}^{-\frac{1}{2}} H^{(l)} W^{(l)}\right),
\end{equation}
where $\tilde{A} = A + I$ denotes the adjacency matrix with added self-loops, $\tilde{D}$ is the degree matrix, $W^{(l)}$ represents the learnable parameters of the $l-th$ layer, $\sigma$ is the activation function.

Furthermore, the Graph Attention Network (GAT) is another important type of GNN, which introduces an attention mechanism to compute the attention weights of nodes with respect to their neighbors.
\begin{equation}
h_i^{(l+1)} = \sigma\left(\sum_{j \in \mathcal{N}(i)} \alpha_{ij} W^{(l)} h_j^{(l)}\right),
\end{equation}
where the attention coefficients are given by \begin{equation}
\alpha_{ij} = \text{softmax}\left(\text{LeakyReLU}(a^T [W h_i || W h_j])\right),
\end{equation}
where $a$  is a learnable attention vector, and $||$  denotes vector concatenation.

As we discussed previously, another mainstream approach for inter-modal integration is the combination of Mobility data. Given that the traffic patterns in mobility data, such as the tidal phenomena during morning and evening rush hours, are essentially spatiotemporal diffusion processes, a graph structure emerges as a natural and optimal choice. It can model the traffic flow propagation between regions through dynamically changing edge weights.
The primary differences among various methods are reflected in their distinct graph construction approaches.

Among them, MV-PN~\cite{mvpnfu2019efficient}, CGAL~\cite{cgalzhang2019unifying}, Region2Vec~\cite{luo2022region2vec}, MVURE
\cite{MVURE}, and ROMER~\cite{romer} all adopt a multi-view graph framework to integrate region information from multiple complementary perspectives.
Specifically, MV-PN~\cite{mvpnfu2019efficient} utilizes an autoencoder to encode the graph network, while CGAL~\cite{cgalzhang2019unifying} employs Graph Convolutional Networks (GCNs) and adversarial networks.
In contrast, Region2Vec~\cite{luo2022region2vec}, MVURE~\cite{MVURE}, and ROMER~\cite{romer} all use Graph Attention Networks (GATs) to adaptively encode region features.
MGFN~\cite{mgfn} considers the temporal dimension during graph construction, where mobility graphs from different time steps are jointly integrated to form a multi-temporal graph.
HREP~\cite{hrep} explores the construction of heterogeneous graphs, in which nodes maintain multiple types of edges to represent diverse relationships.

\item
\textbf{3) \textit{Others.}}
In addition to the aforementioned two classical representation learning methods, there are also some other modeling approaches that, while fewer in number, remain important.
ReFound~\cite{xiao2024refound} employs a \textbf{modality-specific mixture of experts} to strengthen modality modeling and leverages knowledge distillation for cross-modal information fusion.
\citet{zhang2025urban} utilizes \textbf{diffusion model} to model the uncertainty in urban indicators.
Both FlexiReg~\cite{sun2025flexireg} and GraphJCL~\cite{zhao2025graphjcl} first encode the input features through a GNN, and then utilize \textbf{attention mechanisms} to perform feature fusion.

\begin{table*}[!t]
\vspace{-1em}
\resizebox{\linewidth}{!}{
\begin{tabular}{llllllc}
\toprule
   & \textbf{Model} & \textbf{Venue} & \textbf{Modality} &\textbf{Coverage}   & \textbf{Downstream Task} &
   \textbf{Code}\\ \midrule

\multirow{16}{*}{\rotatebox{90}{Single View}}    
& Loc2Vec~\citep{spruyt2018loc2vec}   & Blog 2018    & OSM        &   Global          & Visualization   &-       \\
& Place2Vec \citep{place2vecyan2017itdl}  & SIGSPATIAL 2017&POI&Las Vegas&\makecell[l]{Hierarchy-based Evaluation $\bigm|$ Binary HIT Evaluation
$\bigm|$ Ranking-based HIT Evaluation
$\bigm|$ \\ Place Type Compression
  $\bigm|$ Place Type Profiles}
&\href{https://github.com/BoYanSTKO/place2vec}{Link}\\
& Translocator \citep{translocator}  & ECCV 2022&Geo-Tagged Imagery&Global&Geo-Localization
&\href{https://github.com/ShramanPramanick/Transformer_Based_Geo-Localization}{Link}\\
\cmidrule(lr){2-7}

& Tile2Vec \citep{jean2019tile2vec}  &AAAI 2019 & Satellite Imagery  &-& \makecell[l]{\text{Land Cover Classification} $\bigm|$ \text{Poverty Prediction} $\bigm|$ \text{Health Index Prediction}} &\href{https://github.com/ermongroup/tile2vec}{Link} \\

& READ \citep{readhan2020lightweight}  &AAAI 2020 & Satellite Imagery  & South Korea  &\text{Population Prediction} $\bigm|$ \text{Age Prediction} $\bigm|$ \text{Household Prediction} $\bigm|$ \text{Income Prediction}&\href{https://github.com/Sungwon-Han/READ}{Link} \\

& \citep{han2020learning} &KDD 2020& Satellite Imagery  & \text{Korea} $\bigm|$ \text{Malawi} $\bigm|$ \text{Vietnam}   &\text{Economic Development Evaluation} $\bigm|$ \text{Economic Visual Interpretation}$\bigm|$ \text{Change Detection}  & \href{https://github.com/Sungwon-Han/urban_score.git}{Link} \\ 
 
& \citep{yeh2020using} &Nat Commun 2020& Satellite Imagery & \makecell[l]{\text{Benin} $\bigm|$ \text{Lesotho} $\bigm|$ \text{Malawi} $\bigm|$ \text{Rwanda} $\bigm|$ \text{Sierra} $\bigm|$ \\ \text{Leone} \text{Senegal} $\bigm|$ \text{Tanzania} $\bigm|$ \text{Zambia}} &  \makecell[l]{\text{Asset Wealth Estimation} $\bigm|$ \text{Social Protection Program}  $\bigm|$ \\ \text{Satellite-Estimated Wealth Distribution}$\bigm|$ \text{Temperature Distribution}} & \href{https://github.com/sustainlabgroup/africa_poverty}{Link} \\ 

& \citep{li2023point} &AAAI 2023& Satellite Imagery & \makecell[l]{\text{Kisumu} $\bigm|$ \text{Malindi} $\bigm|$ \text{Nakuru} $\bigm|$ \text{Kenya}} 
& \makecell[l]{Poverty Prediction}
& -\\
 
& HDGE~\citep{hdgewang2017region}&CIKM 2017& Mobility  & Chicago  &
\text{Crime Prediction} $\bigm|$ \text{Income Prediction} $\bigm|$ \text{House Price Prediction}
 &  - \\

& ZE-Mob~\citep{zemobyao2018representing} & IJCAI 2018& Mobility  & New York & \makecell[l]{Functional Region Clssification} & -\\

&MGFN~\citep{mgfn}&IJCAI 2022 & Mobility  
& Beijing 
&\text{Predicting Willingness to Pay} $\bigm|$ \text{Spotting Vibrant Urban Communities} & - \\

&SceneParse~\citep{sceneparselee2021predicting}&AAAI 2021 & Geotagged Imagery & \text{India} $\bigm|$ \text{Kenya} & \text{Poverty Prediction} $\bigm|$ \text{Population Prediction} $\bigm|$ \text{Women's BMI Prediction}
 & \href{https://github.com/sustainlab-group/mapillarygcn}{Link}\\ 

& HGI~\citep{hgihuang2023learning}&ISPRS 2023&POI 
&\text{Shenzhen} $\bigm|$ \text{Xiamen}
& \text{Urban Functional Distributions} $\bigm|$ \text{Population Density Prediction} $\bigm|$ \text{House Price Prediction}
&\href{https://github.com/RightBank/HGI}{Link}\\

&MTE~\citep{mtezhang2024urban}&GISRS 2024&Trajectory
& Shenzhen
& \text{Similar Location Search} $\bigm|$ \text{Land Use Classification} $\bigm|$ \text{Population Density Estimation}
&\href{https://github.com/ZYuSdu/MTE}{Link}\\

& CityFM~\citep{cityfmbalsebre2024} & CIKM 2024 
& OSM
& \text{Singapore} $\bigm|$ \text{Seattle} $\bigm|$ \text{New York}
& \text{Traffic Speed Inference} $\bigm|$ \text{Building Functionality Classification} $\bigm|$ \text{Population Density Estimation}
& - \\
\midrule
\midrule

\multirow{39}{*}{\rotatebox{90}{Dual View}}  
& Geo-SSL~\citep{geosslayush2021geography} & ICCV 2021&Location + Satellite Imagery
& Global (\text{Europe} $\bigm|$ \text{America})
& \makecell[l]{Geotagged Image Classification }
& \href{https://github.com/sustainlab-group/geography-aware-ssl}{Link}
\\

& CSP~\citep{mai2023csp}
&ICML 2023 &Location + Satellite Imagery
& \makecell[l]{\text{New York} $\bigm|$ \text{Tokyo} $\bigm|$ \text{Jakart} $\bigm|$ \text{Beijing}} 
& \text{Location Classification} $\bigm|$ \text{Location Visitor Flow Prediction} $\bigm|$ \text{Next Location Prediction}
&\href{https://github.com/gengchenmai/csp}{Link}\\ 

& SatCLIP~\citep{satclipklemmer2023}& AAAI 2025&Location + Satellite Imagery
& Global
& \makecell[l]{\text{Regression: Air Temperature, Elevation, Median Income, California Housing, Population, Density} $\bigm|$ \\ \text{Classification: Countries, iNaturalist, Biome, Ecoregions}}
&\href{https://github.com/microsoft/satclip}{Link}\\  

& SatCLE~\citep{satclehao2025nature}& WWW 2025&Location + Satellite Imagery
& Global
& \makecell[l]{\text{Regression: Population, Elevation, Carbon Emission} $\bigm|$ \text{Classification: Countries, Land Vegetation}}
&\href{https://github.com/CityMind-Lab/SatCLE}{Link}\\  
& RANGE~\citep{dhakal2025range}& CVPR 2025&Location + Satellite Imagery
& Global
& \makecell[l]{\text{Regression: Air-temperature, Elevation, Population,  Housing-price, Species} $\bigm|$ \\ \text{Classification: Biome, Ecoregion, Country ID}}
&\href{https://github.com/mvrl/RANGE}{Link}\\ 
& TorchSpatial~\citep{wu2024torchspatial}& NeurIPS 2024&Location + Satellite/Geo-Tagged Imagery
& Global
& \makecell[l]{\text{Geo-aware image classification} $\bigm|$ \text{Geo-aware image regression }}
&\href{https://github.com/seai-lab/TorchSpatial}{Link}\\ 
& GPS2Vec~\citep{yin2021gps2vec} & IEEE TMM 2021&Location + Geo-Tagged Imagery 
& Global
& \makecell[l]{ \text{Venue Semantic Annotation} $\bigm|$
\text{Geotagged Image Classification} $\bigm|$ \text{Next Location Prediction}}
& -
\\ 

& GPS2Vec+~\citep{yin2021learning}&ACM MM 2021&Location + Geo-Tagged Imagery
& Global
& \makecell[l]{ \text{Venue Semantic Annotation} 
$\bigm|$
\text{Geotagged Image Classification}}
& \href{https://github.com/yifangyin/GPS2Vec}{Link}
\\ 
& Sphere2Vec~\citep{mai2022sphere2vec}&ISPRS 2023&Location + Geo-Tagged Imagery
& Global 
& \makecell[l]{Geotagged Image Classification}
& \href{https://github.com/gengchenmai/sphere2vec}{Link}\\ 

& GeoCLIP~\citep{vivanco2024geoclip}&NeurIPS 2023&Location + Geo-Tagged Imagery
& Global
& \makecell[l]{Geo-Localization}
&\href{https://github.com/VicenteVivan/geo-clip}{Link}\\                  

& GeoLLM~\citep{manvi2023geollm}&ICLR 2024&Location + Text
& Global
& \makecell[l]{Population Prediction$\bigm|$  \text{Asset Wealth prediction} $\bigm|$ \text{Women Edu Prediction} $\bigm|$ \text{Sanitation Prediction} $\bigm|$ \\ \text{Women BMI Prediction} $\bigm|$ \text{Population Prediction} $\bigm|$ \text{Income prediction} $\bigm|$ \text{Hispanic Ratio Prediction} $\bigm|$ \\ \text{Home Value Prediction}}
&\href{https://github.com/rohinmanvi/GeoLLM}{Link}\\ 

& LLMGeovec~\citep{LLMGeoveche2024geolocation} & AAAI 2025&OSM + Text
& Global
& \makecell[l]{\text{Geographic Prediction} $\bigm|$ \text{Long-term Time series Forecasting} $\bigm|$ \\ \text{Graph-based Spatio-Temporal Forecasting}}
&-\\

& TALE~\citep{talewan2021pre} & TKDE 2022&Location + Trajectory
& \makecell[l]{\text{New York} $\bigm|$ \text{Tokyo} $\bigm|$ \text{Jakart} $\bigm|$ \text{Beijing}} 
& \makecell[l]{Location Classification $\bigm|$ \text{Location Visitor Flow Prediction} $\bigm|$ \text{Next Location Prediction}}
&\href{https://github.com/Logan-Lin/TALE}{Link}\\

& \citep{park2023pre}&ECML-PKDD 2023 &Location + Trajectory
& Global
& \makecell[l]{Next Location Prediction $\bigm|$ \text{Land Usage Classification} $\bigm|$ \text{Transportation Mode Classification}}
&\href{https://github.com/cpark88/ECML-PKDD2023}{Link}\\

& MGeo~\citep{ding2023mgeo}&SIGIR 2023 &Location + Geographic Context
& Hangzhou
& \makecell[l]{Query-POI Matching $\bigm|$ \text
{Ranking task} $\bigm|$ \text{Retrieval task }}
&\href{https://github.com/PhantomGrapes/Mgeo}{Link}\\

& \citep{pan2025largescalegeolocalization}&IEEE TGRS 2025 &Location + Satellite Imagery
& Global
& Geo-Localization
&-\\

\cmidrule(lr){2-7}

& \citep{du2019beyond}&ICDM 2019&POI + Mobility
& Beijing
& \makecell[l]{House Sale Amount Prediction}
& -\\ 

& MV-PN~\citep{mvpnfu2019efficient}&AAAI 2019&POI + Mobility
& Beijing
& \makecell[l]{Regional Mobility Popularity}
&\href{https://github.com/lslrh/multi-view}{Link}\\ 

& CGAL~\citep{cgalzhang2019unifying}&KDD 2019 &POI + Mobility
& Beijing
& \makecell[l]{Regional Mobility Popularity}
&-\\ 

& Region2Vec~\citep{luo2022region2vec}&CIKM 2022&POI + Mobility
& New York
& \makecell[l]{Region Clustering $\bigm|$ \text{Popularity Prediction} $\bigm|$ \text{Crime Prediction}}
&-\\

& MVURE~\citep{MVURE} &IJCAI 2020&POI + Mobility&New York&\makecell[l]{Land Usage Classification $\bigm|$ \text{Crime Prediction}}&\href{https://github.com/mingyangzhang/mv-region-embedding}{Link}\\

& ROMER~\citep{romer} &CIKM 2023&POI + Mobility&New York&\makecell[l]{Land Usage Classification $\bigm|$ \text{Check-in Prediction}}&-\\ 

& HREP~\citep{hrep} & AAAI 2023&POI + Mobility
& New York 
& \makecell[l]{Land Use Classification $\bigm|$ \text{Crime Prediction}}

&-\\

& EUPAS~\citep{chan2024enhanced} & IEEE TIFS 2025&POI + Mobility
& New York
& \makecell[l]{Check-in Prediction $\bigm|$ \text{Crime Prediction} $\bigm|$ \text{Land Usage Classification}}
&-\\ 

& RECP~\citep{recpli2024urban}&AAAI 2024&POI + Mobility
& New York
& \makecell[l]{Land Usage Clustering $\bigm|$ \text{Region Popularity Prediction}}

&-\\  

& ReMVC~\citep{zhang2022region}& TKDE 2022&POI + Mobility
& New York
& \makecell[l]{Land Usage Clustering $\bigm|$ \text{Region Popularity Prediction}}
&-\\ 

& URGent~\citep{urgenthou2022urban} &IEEE TCSS 2022&POI + Mobility
& \text{Beijing} $\bigm|$ \text{Hangzhou}
& \makecell[l]{Traffic Prediction}
&-\\ 

& RegionDCL~\citep{li2023urbanosm} & KDD 2023&POI + OSM
& \text{Singapore} $\bigm|$ \text{New York}
& \makecell[l]{Land Use Prediction $\bigm|$ Population Density Estimation}
&\href{https://github.com/LightChaser666/RegionDCL}{Link}\\

& PG-SimCLR~\citep{pgsimclrxi2022beyond}&WWW 2022 &Satellite Imagery + POI
& Beijing
& \makecell[l]{Region Similarity Analysis $\bigm|$ \text{Socio-Economic Prediction}}
&\href{https://github.com/axin1301/satellite-imagery-POI}{Link}\\ 

& MMGR~\citep{mmgrbai2023geographic} &ISPRS 2023&Satellite Imagery + POI&Shanghai $\bigm|$ Wuhan &\makecell[l]{Urban Function Mapping  $\bigm|$ \text{Population Prediction} $\bigm|$ \text{GDP Prediction}}&\href{https://github.com/bailubin/MMGR.git}{Link}\\ 

& ReFound~\citep{xiao2024refound}&KDD 2024 &Satellite Imagery + POI
& \makecell[l]{\text{Beijing} $\bigm|$ \text{Shanghai} $\bigm|$ \text{Guangzhou} \\ 
\text{Suzhou} $\bigm|$ \text{Shenzhen} } 
& \makecell[l]{Urban Village Detection  $\bigm|$ \text{Commercial Activeness Prediction} $\bigm|$ \text{Population Prediction}}
&-\\ 

& GeoHG~\citep{geohgzou2024learning}& SIGSPATIAL 2025&Satellite Imagery + POI
& \text{Beijing} $\bigm|$ \text{Shanghai} $\bigm|$ \text{Guangzhou} $\bigm|$ \text{Shenzhen}
&\makecell[l]{\text{Carbon Prediction} $\bigm|$ \text{GDP Prediction} $\bigm|$ \text{Population Prediction} $\bigm|$ \\ \text{NightLight Prediction} $\bigm|$ \text{PM2.5 Prediction}}
&-\\

& UrbanCLIP~\citep{urbanclip}& WWW 2024&Satellite Imagery + Text
&\text{Beijing} $\bigm|$ \text{Shanghai} $\bigm|$ \text{Guangzhou} $\bigm|$ \text{Shenzhen}  
&\text{Carbon Prediction} $\bigm|$ \text{GDP Prediction} $\bigm|$ \text{Population Prediction}
&\href{https://github.com/StupidBuluchacha/UrbanCLIP}{Link}\\  

& Urban2Vec~\citep{urban2vec}&AAAI 2020&Street-view Imagery + POI 
& \text{Bay Area} $\bigm|$ \text{Chicago} $\bigm|$ \text{New York}
& \makecell[l]{Income Prediction $\bigm|$ 
 \text{Education Prediction} $\bigm|$  \text{Recial Diversity Prediction}}
& \href{https://github.com/wangzhecheng/urban2vec_.git}{Link}
\\

& USPM~\citep{uspmchen2024profiling}&KDD 2024&Street-view Imagery + Text
& Wuhan
& \makecell[l]{Street Function Prediction $\bigm|$ \text{Socioeconomic Indicator Prediction}}
&-\\

& \citep{li2022predicting} & CIKM 2022&Satellite Imagery + Street-view Imagery
& Beijing
& \makecell[l]{POIs Count $\bigm|$  \text{Commercial Activeness} $\bigm|$  \text{Resident Consumption Population} $\bigm|$ \text{Economic Activity}}
& - 
\\
& \citep{zhang2025urban} &arxiv 2025&POI + Mobility&\text{Manhattan} $\bigm|$ \text{Chicago}&\makecell[l]{House Price Prediction $\bigm|$ \text{Traffic Accident Prediction}$\bigm|$ \text{Carbon Emission Prediction}}
& \href{https://anonymous.4open.science/r/Urban-Incontext-Learning-546B}{Link}
\\
\midrule
\midrule

\multirow{34}{*}{\rotatebox{90}{Multiple View}}

& UrbanFusion~\citep{muhlematter2025urbanfusion} & arxiv 2025 & \makecell[l]{Location + Satellite Imagery \\+ Street-view Imagery + POI \\+ Cartographic Basemaps}
& 56 Cities
& \makecell[l]{House Price Prediction $\bigm|$ \text{Energy Consumption Prediction} $\bigm|$ \text{Crime Prediction} $\bigm|$ \\ \text{Postal code-level Health, Socioeconomic, and Environmental Indicators Prediction} $\bigm|$ \\ \text{Urban Perception} $\bigm|$ \text{ Land
Cover Prediction} $\bigm|$ \text{Coarse-to-Fine Land Use Classification} }
&\href{https://github.com/DominikM198/UrbanFusion}{Link}\\
& AETHER~\citep{aether} & arxiv 2025 & \makecell[l]{Earth Observation Data + POI + Text}
& Greater London &\makecell[l]{Land-Use Classification $\bigm|$ \text{Socioeconomic Distribution Mapping}}&\href{https://github.com/inwind0212/AETHER}{Link}
\\
& GT-Loc~\citep{shatwell2025gtloc} & ICCV 2025 & \makecell[l]{Location + Geo-Tagged Imagery + Time}
& Global &\makecell[l]{Time-of-capture Prediction $\bigm|$ \text{Geo-localization} $\bigm|$ \text{Compositional Retrieval} $\bigm|$ \text{Text-based Retrieval}}&-
\\
& GAIR~\citep{liu2025gair} & arxiv 2025 & \makecell[l]{Location + Satellite Imagery \\+ Street-view Imagery}
& Global &\makecell[l]{Socio-economic Indicator Regression $\bigm|$ \text{Human Perception Regression} $\bigm|$ \text{View Direction Classification} $\bigm|$ \\ \text{Imaging Platform Classification}$\bigm|$ \text{Burn Scar Segmentation}$\bigm|$\text{Crop Type Mapping} $\bigm|$ \\ \text{Cropland Polygon Segmentation}$\bigm|$ \text{Geo-aware Image Regression} $\bigm|$ \text{Species Recognition} $\bigm|$ \\ \text{Flickr Image Classification} }&-
\\
& GeoDecoder~\citep{geodecoder} & CVPR 2023 & \makecell[l]{Location + Geo-Tagged Imagery\\ + Scene Labels}
& Global &Geo-Localization&- \\
& PIGEON~\citep{haas2024pigeon} & CVPR 2024 & \makecell[l]{Location + Street-view Imagery + Text}
& Global &Geo-Localization&\href{https://github.com/LukasHaas/PIGEON}{Link} \\
& CityGuessr~\citep{kulkarni2024cityguessr} & ECCV 2024 & \makecell[l]{Address + Video + Scene Labels}
& Global &Geo-Localization&- \\
& G3~\citep{jia2024g3} & NeurIPS 2024 & \makecell[l]{Location + Geo-Tagged Imagery + Text}
& Global &Geo-Localization&\href{https://github.com/Applied-Machine-Learning-Lab/G3}{Link} \\
& Georanker~\citep{georanker} & NeurIPS 2025 & \makecell[l]{Location + Geo-Tagged Imagery + Text}
& Global &Geo-Localization&\href{https://github.com/Applied-Machine-Learning-Lab/GeoRanker}{Link} \\

&\citep{jia2025towardsinterpretable} & arxiv 2025 & \makecell[l]{Location + Geo-Tagged Imagery + Text}
& Global &Geo-Localization&- \\
&\citep{lindenberger2025scalingimagegeolocalization} & NeurIPS 2025 & \makecell[l]{Location + Satellite Imagery + Street-view Imagery}
& \makecell[l]{BEDENL $\bigm|$ EuropeWest $\bigm|$ UK+IE} &Geo-Localization&\href{https://scaling-geoloc.github.io/}{Link} \\
& GLOBE~\citep{globe} & NeurIPS 2025 & \makecell[l]{Location + Geo-Tagged Imagery + Text}
& Global &Geo-Localization&\href{https://github.com/lingli1996/globe}{Link} \\
\cmidrule(lr){2-7}
\cmidrule(lr){2-7}
& RegionEncoder~\citep{regionencoderjenkins2019unsupervised} & CIKM 2019 & Satellite Imagery + POI + Mobility
& \text{Chicago} $\bigm|$ \text{New York}
& \makecell[l]{House Sale Prediction}
&\href{https://github.com/porterjenkins/region-encoder}{Link}\\
& M3G~\citep{m3ghuang2021learning} & AAAI 2021& Street-view Imagery + POI + Mobility
& \text{Chicago} $\bigm|$ \text{New York}
& \makecell[l]{Crime Prediction $\bigm|$ \text{Bike Flow Prediction} $\bigm|$  \text{Average Personal Income Prediction}}
&\href{https://github.com/tianyuanhuang/M3G}{Link}\\

& Geo-Tile2Vec~\citep{geotile2vecluo2023geo}&ACM TSAS 2023& Street-view Imagery + POI + Mobility
& \text{Beijing} $\bigm|$ \text{Nanjing} $\bigm|$ \text{Nanchang}
& \makecell[l]{Land Use Classification $\bigm|$ \text{POIs Classification} $\bigm|$ \text{Restaurant Average Price Prediction}}
&-\\

& MuseCL~\citep{yong2024musecl}&IJCAI 2024& \makecell[l]{Satellite Imagery + Street-view Imagery \\ + POI + Mobility}
& \text{Beijing} $\bigm|$ \text{Shanghai} $\bigm|$ \text{New York}
& \makecell[l]{Land Usage Clustering $\bigm|$ \text{Popularity Prediction}}
&\href{https://github.com/XixianYong/MuseCL}{Link}\\

& UrbanVLP~\citep{hao2024urbanvlp}&AAAI 2025&Satellite Imagery + Street-view Imagery + Text
& \text{Beijing} $\bigm|$ \text{Shanghai} $\bigm|$ \text{Guangzhou} $\bigm|$ \text{Shenzhen}
&\makecell[l]{\text{Carbon Prediction} $\bigm|$ \text{GDP Prediction} $\bigm|$ \text{Population Prediction}  $\bigm|$  \text{NightLight Prediction} $\bigm|$ \\ \text{House Price Prediction} $\bigm|$ \text{POI Prediction}}
&\href{https://github.com/CityMind-Lab/UrbanVLP}{Link}\\

& KnowCL~\citep{2023KnowCL}&WWW 2023&Satellite Imagery + Knowledge Graph
& \text{Beijing} $\bigm|$ \text{Shanghai} $\bigm|$ \text{New York}
& \makecell[l]{Population Prediction $\bigm|$ \text{Economy Prediction} $\bigm|$ \text{Crime Prediction}}
&\href{https://github.com/quanweiliu/KnowCL}{Link}
\\

& HAFusion~\citep{hafusionsun2024urban}&ICDE 2024& POI + Mobility + Land Usage
& \text{New York} $\bigm|$ \text{Chicago} $\bigm|$ \text{San Francisco}
& \makecell[l]{Crime Prediction $\bigm|$ \text{Check-in Prediction}}
&\href{https://github.com/matt8707/ha-fusion}{Link}\\

& HUGAT~\citep{kim2022hugat}&IEEE Access 2025&POI + Land Usage+ Check-in + Taxi Record
& New York
& \makecell[l]{Crime Prediction $\bigm|$ \text{Check-in Prediction}}
&-\\
& FlexiReg~\citep{sun2025flexireg}&KDD 2025&\makecell[l]{Satellite Imagery + Street-view Imagery + POI \\+ Text + Land Use +
Geographic Neighbor}&\makecell[l]{\text{New York} $\bigm|$ \text{Chicago} $\bigm|$ \text{San Francisco} $\bigm|$ \text{Singapore} $\bigm|$ \\ \text{Lisbon}} &\makecell[l]{Population Prediction $\bigm|$ \text{Check-in Prediction} $\bigm|$ \text{Crime Prediction} $\bigm|$ \text{Service Call Prediction}}&-\\
&GURPP~\cite{gurpp} &KDD 2025&Satellite Imagery + POI + Mobility&\text{New York} $\bigm|$ \text{Chicago}&\makecell[l]{Crash Prediction $\bigm|$ \text{Check-in Prediction} $\bigm|$ \text{Crime Prediction}}&\href{https://doi.org/10.5281/zenodo.15565147}{Link}\\
&~\cite{namgung2025lessismore}&arxiv 2025&\makecell[l]{Satellite Imagery + Building Footprints \\+ POI + AOI}&\text{New York} $\bigm|$ \text{Delhi}&\makecell[l]{Population Prediction $\bigm|$ \text{Crime Prediction} $\bigm|$ \text{Greenness Score Prediction $\bigm|$ \text{Land Use Classification}} }&\href{https://github.com/MinNamgung/CooKIE}{Link}\\
&\cite{zhao2025modality} &ECML-PKDD 2025&\makecell[l]{Satellite Imagery + Street View Imagery + POI\\+ Taxi Flow + Road Network} &\text{New York} $\bigm|$ \text{Chicago}&\makecell[l]{Crime Prediction $\bigm|$ \text{Check-in Prediction} $\bigm|$ \text{Traffic Crash Prediction}}&- \\
&GraphJCL~\cite{zhao2025graphjcl} &ECML-PKDD 2025&\makecell[l]{Satellite Imagery + Street View Imagery + POI \\ + Taxi Flow + Road Network} &\text{New York} $\bigm|$ \text{Chicago}&\makecell[l]{Crime Prediction $\bigm|$ \text{Check-in Prediction} $\bigm|$ \text{Traffic Crash Prediction}}&- \\
&\cite{arcl}&SIGSPATIAL 2025&POI + Mobility + Land Use&\text{New York} $\bigm|$ \text{Chicago} $\bigm|$ \text{San Francisco}&\makecell[l]{Crime Prediction $\bigm|$ \text{Check-in Prediction} $\bigm|$ \text{Service Call Prediction}} & \href{https://github.com/Longsuni/ARC}{Link} \\
&MobCLIP~\cite{wen2025mobclip}&arxiv 2025&\makecell[l]{Satellite Imagery + POI + Mobility \\+ \text{Demographic Statistics}} &China (Nationwide)&\makecell[l]{Population Prediction $\bigm|$ \text{Elderly Population Ratio Prediction} $\bigm|$ \text{Hukou Separation Rate Prediction} $\bigm|$ \\ \text{Crime Prediction}$\bigm|$ \text{Nighttime Light Prediction}$\bigm|$\text{Per Capita Housing Area Prediction} $\bigm|$ \\ \text{Energy Consumption Prediction}$\bigm|$ \text{Elevation Prediction} $\bigm|$ \text{Offline Consumption Amount Prediction} $\bigm|$ \\ \text{Forest Coverage Prediction} }&\href{https://github.com/ylzhouchris/MoRA}{Link} \\
&MGRL4RE~\cite{MGRL4RE} &ACM TIST 2025&\makecell[l]{POI + Mobility + Region boundary geometry} &\text{Manhattan} &\makecell[l]{\text{Popularity Prediction} $\bigm|$ \text{Crime Prediction} $\bigm|$ \text{Land-use clustering}}&- \\
&UrbanLN~\cite{urbanln} &AAAI 2026&\makecell[l]{Satellite Imagery + Street View Imagery + Text} &\text{Beijing} $\bigm|$ \text{Shanghai} $\bigm|$ \text{Shenzhen} $\bigm|$ \text{New York}&\makecell[l]{\text{Population Prediction} $\bigm|$ \text{GDP Prediction} $\bigm|$ \text{NightLight Prediction}  $\bigm|$  \text{Carbon Prediction} $\bigm|$ \\ \text{Restaurant Comments Prediction} $\bigm|$ \text{POI Prediction} $\bigm|$ \text{Crime Prediction}}&\href{https://github.com/YimeiZhang0229/UrbanLN}{Link} \\
&UrbanMMCL~\cite{UrbanMMCL} &ISPRS 2026&\makecell[l]{Satellite Imagery + Street View Imagery + Text \\+ \text{POI + Mobility}} &\text{Shenzhen}&\makecell[l]{\text{Population Prediction} $\bigm|$ \text{Pollutant Concentration Prediction} $\bigm|$ \text{Land Use Classification}}&- \\

\midrule
\end{tabular}
}
\caption{A summary of deep learning-based works in geospatial representation learning.}
\label{tab:summarizations}
\end{table*}

\end{itemize}

\vspace{-1.5em}
\section{Application Perspective}
\label{sec:application}

\subsection{Socioeconomic Indicator  Prediction}
Regional indicators include various statistical measures of a geospatial area from both environmental and social perspectives, such as \textit{regional GDP and poverty}~\citep{jean2019tile2vec,sheehan2019predicting,urban2vec,yeh2020using,yin2021learning,li2023point,hao2024urbanvlp,geohgzou2024learning}, \textit{crime}~\citep{hdgewang2017region}, \textit{mobility popularity}~\citep{regionencoderjenkins2019unsupervised,cgalzhang2019unifying,liu2020learning,yong2024musecl}, \textit{house prices}~\citep{regionencoderjenkins2019unsupervised,du2019beyond,yong2024musecl}, \textit{population}~\citep{sceneparselee2021predicting,li2022predicting,fan2023urban,yong2024musecl,hao2024urbanvlp,geohgzou2024learning}, \textit{energy}~\cite{prabowo2023continually}, and \textit{air quality}~\citep{hao2024urbanvlp,geohgzou2024learning,shao2022long}. 
Traditionally, data collection has relied on extensive and costly field research~\citep{mascarenhas2010role,zou2025deep}, such as population censuses and air quality monitoring. However, limited human resources and budgets often hinder comprehensive data collection~\citep{force1997human,ringold2013data}, prompting the development of statistical methods
to estimate regional indicators and enhance accuracy.

At early stages, the application of geospatial representation was primarily limited to urban scales and a narrow range of indicators, due to the constraints of representation models and the availability of real-world datasets. Most studies initially focused on the prediction of regional \textbf{crime rates} and \textbf{house prices}. HDGE~\citep{hdgewang2017region} utilized taxi mobility data to estimate these rates, while RegionEncoder~\citep{regionencoderjenkins2019unsupervised} enhanced model performance by integrating regional satellite imagery and POI data. Additionally, the multi-view graph structure proposed by \citet{du2019beyond} improved the understanding of topical relationships between regions, resulting in better predictions of house sales. These approaches have effectively provided insights for urban planning and real estate investment.

With the advancement of dual view representation, applications have expanded to include other indicators like \textbf{popularity} prediction, quantified by regional check-in counts~\citep{cgalzhang2019unifying,liu2020learning}, and the prediction of additional economic indicators~\citep{sheehan2019predicting,urban2vec,yeh2020using,yin2021learning,li2023point}.
To address diverse practical demands, geospatial representation in socioeconomic indicator prediction has evolved to encompass broader scales and tasks. Recent efforts, such as those by \citet{sceneparselee2021predicting,li2022predicting,fan2023urban}, have demonstrated the potential for geospatial representation in predicting \textbf{poverty} and \textbf{population density} on a wider or even global scale. The fusion of multi-modal data and advanced representation learning has significantly contributed to improvements in generalization across applications. For instance, MuseCL~\citep{yong2024musecl} employs contrastive learning to combine features from satellite imagery, street-view imagery, POI, and mobility data, enhancing prediction accuracy. This contrastive learning approach has been utilized in several recent studies~\citep{urbanclip,hao2024urbanvlp,geohgzou2024learning,mtezhang2024urban,xiao2024refound}, improving accuracy in predictions of various regional indicators \textit{(e.g., GDP, air quality, night light)}~\citep{geohgzou2024learning,mtezhang2024urban}. SatCLIP~\citep{satclipklemmer2023} introduces a global location embedding to predict air temperature and population density at a global scale, while GeoLLM~\citep{manvi2023geollm} further enhances the performance of global location embeddings by incorporating large language models (LLMs).

\begin{figure}[t]
    \centering
    \includegraphics[width=1.0\linewidth]{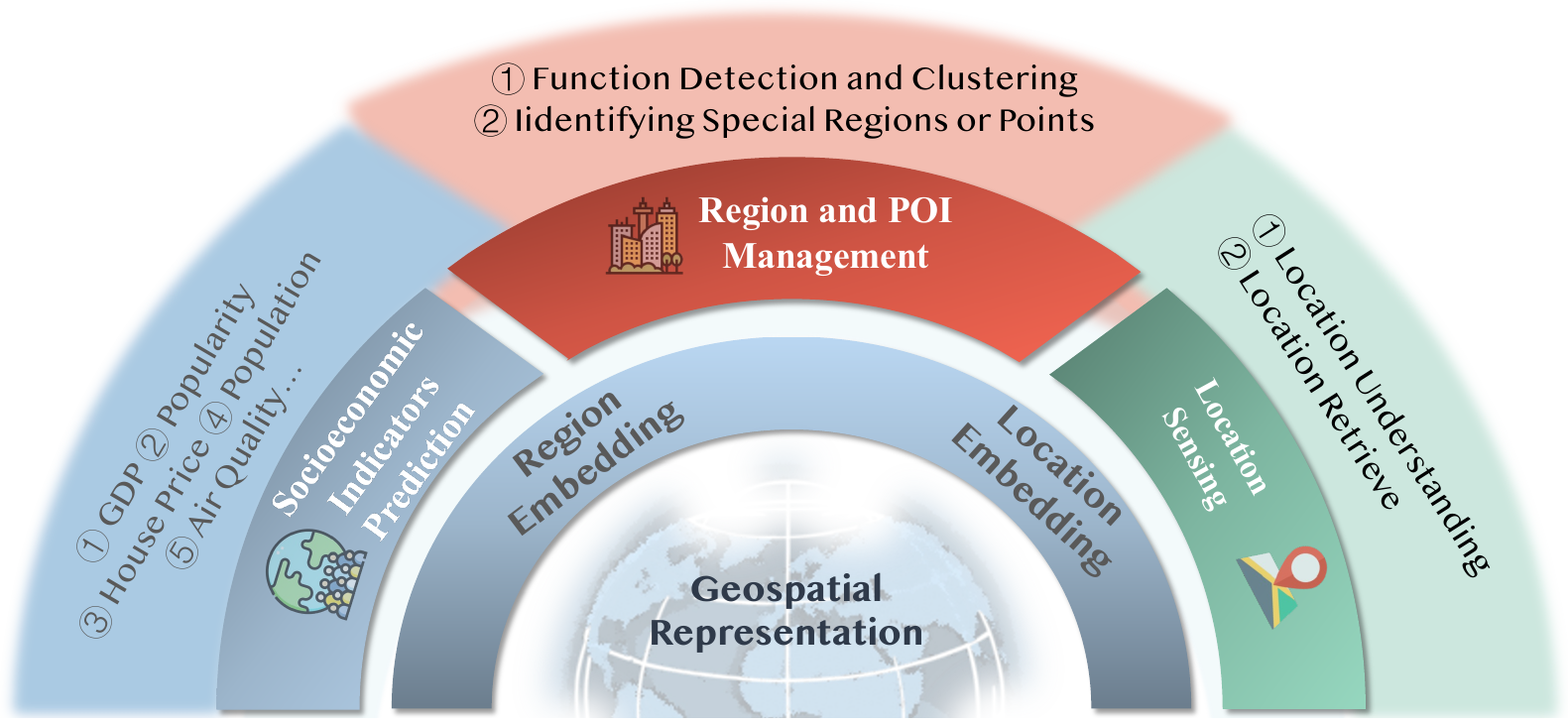}
    \vspace{-1.5em}
    \caption{Taxonomy of Application for Geospatial Embedding.}
    \label{fig:app_location}
    \vspace{-1.5em}
\end{figure}

\vspace{-0.2em}
\subsection{Region and POI Management}
The rapid dynamics of human activities render administrative boundaries and other manually designed boundaries insufficient for meeting the real-world requirements of public services. The functional similarities and socio-economic connections between regions and locations are challenging to detect and quantify using traditional methods~\citep{zou2025deep}. Location and region embeddings can facilitate the detection and management of POI and regional characteristics through a data-driven approach, such as automatically clustering regions into different functional groups~\citep{zemobyao2018representing,zhang2021multi,luo2022region2vec,wu2022multi} and identifying special regions or locations~\citep{wang2018learning,mtezhang2024urban,place2vecyan2017itdl,spruyt2018loc2vec}.

ZE-Mob~\citep{zemobyao2018representing} utilizes taxi mobility data to identify the urban functions of regions in New York City. Similarly, \citet{wang2018learning} employs the same type of data to detect popular zones in urban communities. Numerous unsupervised region embedding methods~\citep{zhang2022region,geotile2vecluo2023geo,fan2023urban,mmgrbai2023geographic,li2024urban} have demonstrated impressive performance in land use clustering and detection. For example, ReFound~\citep{xiao2024refound} introduces a contrastive learning-based framework that efficiently detects urban villages in cities through satellite imagery and POIs. 

\vspace{-0.6em}
\subsection{Location Sensing}

Geospatial representation learning essentially provides comprehensive information about geospace and contribute to advanced geospatial analytical applications like \textbf{location understanding}~\citep{talewan2021pre,park2023pre} and \textbf{geo-localization}~\citep{pramanick2022world}, as demonstrated in Figure \ref{fig:app_location}. Location understanding helps to gain more information from geospatial perspectives to enhance the useful features for various GeoAI tasks. For example, \citet{talewan2021pre} enhances the representation of traffic trajectories with the pre-trained location embedding to improve the model's performance on visitor flow prediction in cities. \citet{park2023pre} introduces location embedding to trajectory analysis to improve prediction accuracy of transportation modes. Many other research like spatio-temporal prediction of urban traffic flow~\cite{urgenthou2022urban,cityfmbalsebre2024}, regional climate variation~\cite{satclehao2025nature,dhakal2025range,yeh2020using} and air quality~\cite{geohgzou2024learning} also utilize geospatial embedding as useful basic information for learning the spatial correlations of various phenomena.

Geo-localization~\cite{vivanco2024geoclip,jia2024g3,georanker,jia2025geoarena,globe,wang2025gre,dou2024gaga,xu2024addressclip,han2025swarm,zheng2025graphgeo} is a task of predicting precise GPS coordinates (latitude and longitude) from image content, placing extremely demanding requirements on the geospatial knowledge and sophisticated reasoning capabilities of models.
Existing practices have transcended conventional classification and image retrieval approaches, transitioning toward geospatial representation-based alignment, retrieval-augmented generation (RAG), and multi-agent collaborative reasoning frameworks.
GeoCLIP~\cite{vivanco2024geoclip} pioneered the formulation of global geo-localization as an image-to-GPS retrieval task, representing the Earth as a continuous function to address the limitations of conventional classification approaches.
G3~\cite{jia2024g3} introduced a Geo-alignment mechanism that employs contrastive learning to jointly align multimodal representations of images, GPS coordinates, and textual descriptions, coupled with a RAG mechanism for generating diversified predictions.
GeoRanker~\cite{georanker} developed a distance-aware ranking framework that leverages LVLMs to model intricate spatial interactions between query images and candidate geographic entities, thus enhancing fine-grained localization precision.
GLOBE~\cite{globe} and GRE Suite~\cite{wang2025gre} focus on leveraging reinforcement learning (Group-relative Policy Optimization) combined with task-specific rewards (such as localizability, visual grounding consistency, and geo-localization accuracy) to enhance the recognition and reasoning capabilities of LVLMs, generating interpretable localization results.

\begin{figure}[h!]
    \centering
    \includegraphics[width=1.0\linewidth]{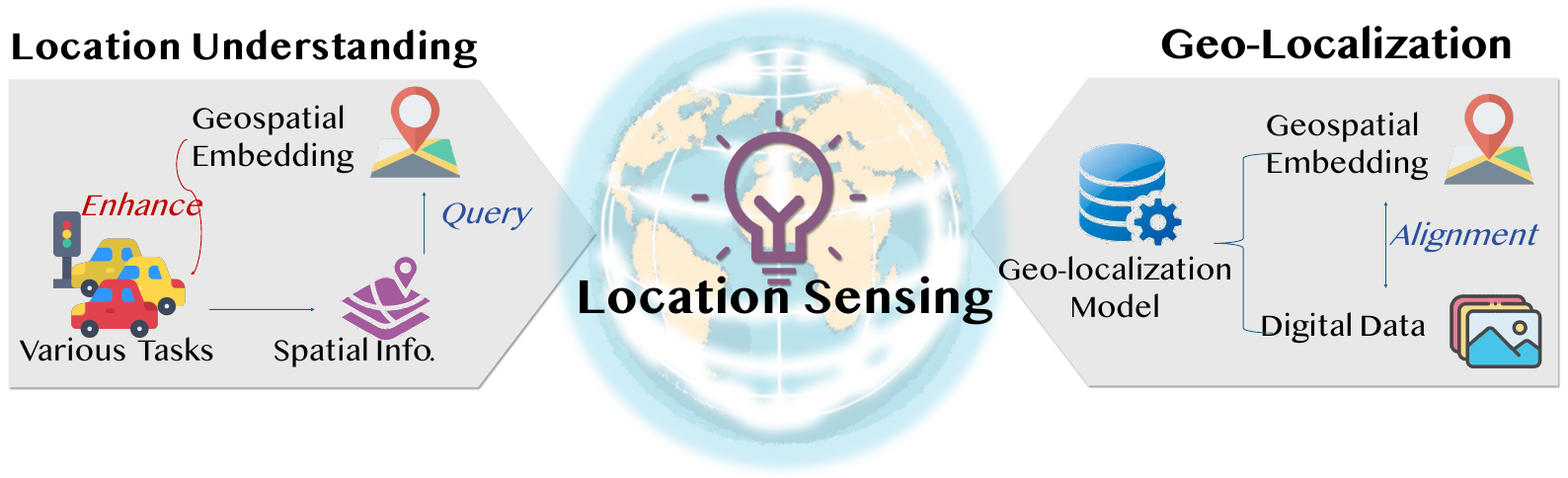}
    \vspace{-1em}
    \caption{Application of Geospatial Embeddings for Location Sensing.}
    \label{fig:app_location}
    \vspace{-1em}
\end{figure}
\vspace{-1em}
\section{Geospatial Representation Learning in the LLM Era}
\label{sec:challenge}
\begin{figure}[h]
    \centering
    \includegraphics[width=1.0\linewidth]{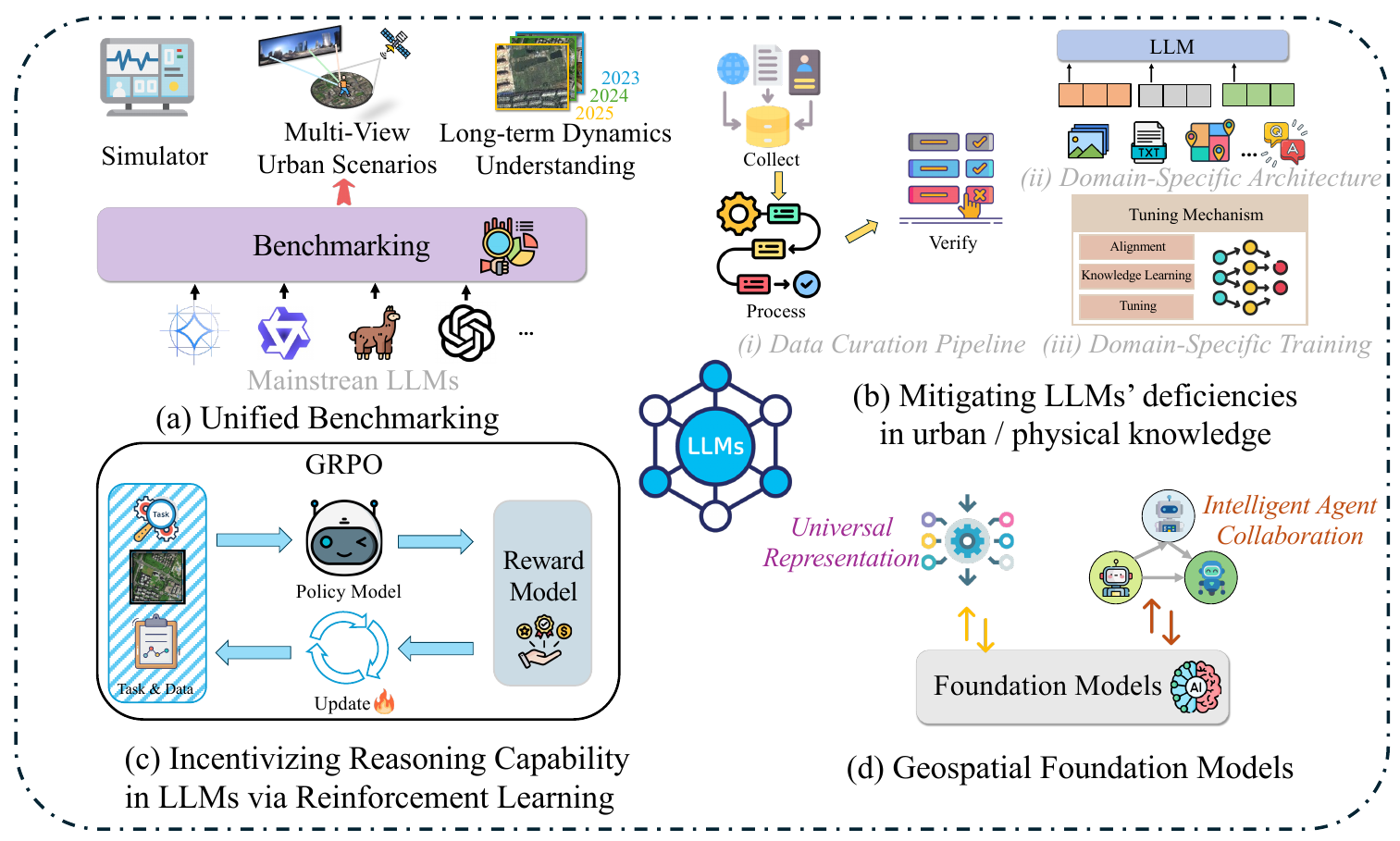}
    \vspace{-1em}
    \caption{Geospatial Representation Learning in the LLM Era.}
    \label{fig:llmera}
    \vspace{-1em}
\end{figure}

In the LLM era, with the rapid advancement of emerging technologies, approaches to geospatial representation learning have undergone significant transformations. As illustrated in Figure~\ref{fig:llmera}, we categorize the development of GRL in the LLM era into four main directions: (1) unified benchmarking; (2) mitigating LLMs' inherent deficiencies in urban and physical world knowledge; (3) incentivizing reasoning capabilities in LLMs via reinforcement learning; and (4) geospatial foundation models. These categories will be elaborated in the subsequent sections.

\subsection{Unified Benchmarking}
In recent years, LLMs~\cite{deepseekai2024deepseekv3technicalreport,comanici2025gemini,bai2025qwen2} have demonstrated robust multimodal understanding capabilities (encompassing text, coordinates, and geographic descriptions)~\cite{yin2024survey,zhang2024mm}, the ability to perform spatial reasoning~\cite{zha2025enable,chen2024spatialvlm}, and proficiency in knowledge integration (including background knowledge of geography, history, culture, etc.)~\cite{an2025towards,zou2025deep}.
In the geospatial domain, the complexity of spatial data—encompassing 3D spatial coordinates, heterogeneous data modalities, and cross-scale variations (from centimeters to kilometers), poses significant challenges to LLMs' capabilities in geospatial applications.
Consequently, the academic community has conducted multi-level benchmarking~\cite{liu2025citylens,zhou2025urbench,feng2025citybench,xuan2025dynamicvl,urbanvideo} of LLMs, aiming to systematically evaluate the capabilities and limitations of large models in processing complex urban data and tasks.

CityBench~\cite{feng2025citybench} presents a systematic evaluation framework built upon \textbf{interactive simulators}, designed to evaluate the performance of LLMs and Vision-Language Models (VLMs) on diverse urban research tasks spanning perception and understanding as well as decision-making.
CityLens~\cite{liu2025citylens} constructs a multimodal benchmark leveraging satellite and street-view imagery to assess LVLMs on \textbf{urban socioeconomic prediction} across 11 indicators and 6 domains, revealing their challenges in numerical estimation and geospatial reasoning.
In terms of \textbf{spatial perspective}, UrBench~\cite{zhou2025urbench} evaluates LMMs across complex multi-view urban scenarios encompassing satellite, street, and cross-view imagery. With 14 task types covering geo-localization, scene reasoning, scene understanding, and object understanding, it exposes notable limitations of current models in processing cross-view relationships and ensuring cross-perspective consistency.
DynamicVL~\cite{xuan2025dynamicvl} focuses on \textbf{long-term urban dynamics understanding} by constructing the DVL-Suite framework, which leverages high-resolution multi-temporal remote sensing imagery spanning nearly two decades. It encompasses six core tasks ranging from pixel-level change detection to region-level dense temporal narratives, aiming to evaluate the limitations of Multimodal Large Language Models (MLLMs) in processing extended time series and conducting quantitative change analysis.

\subsection{Integrating LLMs with real-world knowledge}
The inherent deficiencies of LLMs in processing urban and physical world knowledge are primarily addressed and enhanced through three dimensions~\cite{feng2025urbanllava,feng2024citygpt,li2024urbangpt,li2024opencity,zhang2025urbanmllm,han2024geosee,kuckreja2023geochat,zhan2025skyeyegpt,nedungadi2024mmearth}: \textbf{data customization}, \textbf{architectural integration}, and \textbf{domain-specific reasoning mechanisms}.

In terms of data and multimodal integration, UrbanMLLM~\cite{zhang2025urbanmllm} achieves complementary learning of visual information from different perspectives by collecting a large-scale cross-view urban imagery corpus (integrating macroscopic remote sensing and microscopic street-view information) and designing a structured interleaved image-text pre-training paradigm. 
UrbanLLaVA~\cite{feng2025urbanllava} takes this further by designing the UData pipeline to uniformly model four major types of urban data (including geospatial, street-view, satellite, and trajectory data), and comprehensively captures the multifaceted nature of urban systems through the construction of multi-perspective instruction datasets (local, trajectory, and global perspectives). CityGPT~\cite{feng2024citygpt}, on the other hand, focuses on constructing the CityInstruction instruction-tuning dataset by simulating human mobility behavior, which incorporates explicit spatial reasoning steps to inject offline urban spatial cognition and experiential knowledge into the model.

At the architecture and encoding level, to overcome visual isolation, UrbanMLLM~\cite{zhang2025urbanmllm} introduces a cross-view perceiver module that explicitly fuses the regional contextual information from satellite imagery with the fine-grained appearance details from street-view imagery through a cross-attention mechanism.
To address the challenges of spatiotemporal prediction tasks, both UrbanGPT~\cite{li2024urbangpt} and OpenCity~\cite{li2024opencity} enhance the spatiotemporal perception capabilities of LLMs by integrating spatiotemporal dependency encoders. OpenCity~\cite{li2024opencity} combines the Transformer architecture with graph neural networks and employs Instance Normalization techniques to tackle the heterogeneity and distribution drift issues in traffic data, achieving robust zero-shot cross-regional generalization capabilities. UrbanGPT~\cite{li2024urbangpt}, meanwhile, utilizes a lightweight alignment module to map spatiotemporal dependency representations into the LLM's hidden space, and encodes temporal and geographical information in the form of structured prompt text to facilitate spatiotemporal context alignment.

In terms of training and inference paradigms, UrbanLLaVA~\cite{feng2025urbanllava} adopts a multi-stage training pipeline that decouples the learning of spatial reasoning capabilities from domain knowledge acquisition, ensuring stable and balanced performance across heterogeneous multimodal tasks. CityGPT~\cite{feng2024citygpt} applies Self-Weighted Fine-Tuning (SWFT), which can automatically assess data quality and adjust loss weights. UrbanGPT~\cite{li2024urbangpt}, on the other hand, adopts an approach of generating predictive tokens followed by outputting final numerical values through a regression layer, addressing the issue of insufficient precision in LLMs for continuous numerical regression tasks.




\subsection{Incentivizing LLM Reasoning via Reinforcement Learning}

LLMs typically require fine-tuning to accommodate domain-specific tasks or achieve alignment with human values. Reinforcement Learning from Human Feedback (RLHF)~\cite{ouyang2022training}, the cornerstone technology powering ChatGPT, incorporates human preferences into model training to ensure outputs better reflect human expectations. Building upon this foundation, DeepSeek-R1~\cite{guo2025deepseekr1} introduced Group Relative Policy Optimization (GRPO), demonstrating that reinforcement learning-based alignment can enhance reasoning capabilities even with constrained training data and computationally efficient reward mechanisms.

This technique has been widely adopted across multiple domains~\cite{huang2025visionr1,luo2025guir1,liu2025timer1}. In geospatial representation learning, the GRPO strategy enables models to acquire geospatial reasoning patterns that are more robust, interpretable, and generalizable compared to supervised fine-tuning (SFT), demonstrating particularly significant advantages when handling unseen regions or out-of-distribution scenarios.
Regarding structured strategies, Geo-R1~\cite{xu2025geor1} employs a two-stage approach: first injecting a geospatial thinking paradigm through SFT to establish scaffolding, followed by GRPO to enhance reasoning quality. Traffic-R1~\cite{zou2025trafficr1} integrates offline RL with human expert knowledge and online open-world RL for autonomous exploration, aiming to generate human-like transparent decision-making through self-iteration. 
CityRiSE~\cite{liu2025cityrise} cultivates transferable reasoning skills by constructing auxiliary perception and general visual reasoning datasets.

In terms of reward design, models guide complex geospatial reasoning by imposing rigorously verifiable rewards on both output results and reasoning processes. 
GLOBE~\cite{globe}, targeting image geo-localization tasks, incorporates multiple reward metrics including Locatability Reward, Visual Grounding Consistency Reward, and Geo-localization Accuracy Reward to directly optimize spatial localization precision. This approach significantly enhances sample efficiency and cross-dataset generalization capability, particularly in few-shot scenarios. 
Meanwhile, Geo-R1~\cite{xu2025geor1} leverages cross-view pairing as a proxy task, rewarding accuracy, length, and repetition to ensure that the model effectively learns and exploits complex visual cues for geo-localization reasoning.


\subsection{Geospatial Foundation Models}
Current practices of geospatial foundation models (GFMs)~\cite{earthai,brown2025alphaearth,Prithvi,szwarcman2024prithviv2,sun2022ringmo,nedungadi2024mmearth} primarily focus on two paradigms: constructing universal, multimodal Earth representations and developing intelligent agent collaboration systems. 
The Earth AI framework~\cite{earthai} adopts a modular integration strategy, orchestrated by a Gemini-powered~\cite{comanici2025gemini} geospatial reasoning agent, aimed at addressing complex, multi-step geospatial queries. The system integrates specialized foundation models across three core domains: imagery, population, and environment.
Targeting Earth Observation, AlphaEarth Foundations~\cite{brown2025alphaearth} adopts an implicit embedding field model approach. Through a Space-Time Precision encoder, it fuses multi-source data—including optical, radar, LiDAR, environmental, and textual information—to generate high-precision, global annual embedding fields at 10-meter resolution. This framework enables continuous temporal support and efficient global mapping.
Another significant pathway to realizing GFM is the Prithvi model family~\cite{Prithvi,szwarcman2024prithviv2}, based on self-supervised learning, which focuses on multi-spectral and multi-temporal remote sensing representation learning. 
Prithvi~\cite{Prithvi} employs a Masked Autoencoder (MAE) approach for large-scale pretraining on HLS multi-spectral satellite imagery, with core modifications involving the use of 3D positional embeddings and 3D patch embeddings to effectively process spatiotemporal data. 
Its successor, Prithvi-EO-2.0~\cite{szwarcman2024prithviv2}, further expands the training dataset to 4.2M global samples and explicitly introduces a unique metadata processing mechanism in the model architecture. This mechanism incorporates temporal (year, day of year) and geolocation (latitude, longitude) metadata as weighted bias terms into the embedding representations, substantially enhancing the model's generalization capability and data efficiency across different resolutions and geographic regions.


\section{Future Directions}
Despite the substantial progress in geospatial representation learning recently, several persistent challenges remain, underscoring critical avenues for future exploration and innovation in the LLM and foundation model era.

\textbf{Standard Datasets, Codebases, and Downstream Tasks.}
In contrast to general domains such as Computer Vision (CV)~\citep{voulodimos2018deep} and Natural Language Processing (NLP)~\citep{min2023recent}, a review of Geospatial Representation Learning reveals significant deficiencies in current research. These include the lack of standardized datasets, codebases, and evaluation metrics for downstream tasks, as shown in Table~\ref{tab:summarizations}. 
The lack of uniformity complicates model comparisons and hampers the quantification of contributions from various data modalities, such as satellite imagery, POIs, mobility data, and so on. 
TorchSpatial~\cite{wu2024torchspatial} takes steps toward this goal, yet its implementation is confined to location encoding.
Additionally, suboptimal open-source practices, with many studies failing to provide comprehensive code and data, limit reproducibility and scalability.
Constructing a unified benchmark encompassing data, models, and downstream tasks is a valuable direction for future development.



\textbf{Investigation of the contribution of different modalities to downstream tasks.}
One of the most prominent characteristics of geospatial representation learning lies in its diverse data sources, including satellite imagery, street-view imagery, POIs, textual data, mobility patterns, and trajectories, among others, as illustrated in Figure~\ref{fig:taxonomy} and Table~\ref{tab:summarizations}. However, while these diverse data modalities have all been demonstrated to be effective, it remains unclear which modality contributes most significantly to a given downstream task and provides the most informative features~\cite{wang2021exploring}. 
Therefore, establishing a fair and systematic comparison framework is essential for advancing this domain. Such investigations are instrumental in revealing the multi-scale and multi-dimensional nature of geospatial phenomena, thereby promoting theoretical innovation at the intersection of geographic information science and machine learning. 
From a practical standpoint, a rigorous understanding of modality-specific contributions can guide data-driven resource allocation strategies, enabling practitioners to prioritize the acquisition and processing of data modalities that contribute most significantly to the target task under budget and data availability constraints.

\textbf{Continuous and Physics-informed Representation Learning.}
Existing frameworks primarily utilize neural networks to extract features from structured grids or graph-based regions. However, these approaches depend heavily on discretizing space into fixed units, which inevitably leads to the Modifiable Areal Unit Problem (MAUP)~\cite{zeng2020revisiting} and resolution dependency~\cite{guo2023enhancing}. 
Future research should transcend these limitations by advancing continuous~\cite{satclehao2025nature} and physics-informed representation learning, such as Implicit Neural Representations (INRs)~\cite{sitzmann2020implicit} or Neural Fields~\cite{ijcai2024p803}, to model geospatial phenomena as continuous and resolution-independent functions. Furthermore, rather than treating geospatial data as generic images or graphs as seen in current deep learning paradigms, future architectures should integrate domain-specific inductive biases. By incorporating principles like Tobler’s First Law of Geography or spatial diffusion equations directly into the loss functions, models can learn representations that are not only statistically powerful but also consistent with the physical and topological laws governing urban dynamics and environmental processes.

\textbf{Mixture of Experts and Agentic Learning.}
Current geospatial representation learning approaches commonly adopt a monolithic architecture, insufficiently capturing the inherent heterogeneity~\cite{li2024opencity} among diverse geographic regions, spatial scales, and data modalities—including disparate landscape patterns, degrees of urbanization, and multimodal data features.
Mixture of Experts (MoE)~\cite{mu2025comprehensive,xiao2024refound} naturally accommodates the heterogeneous nature of geospatial data through expert specialization, allowing dedicated experts to be assigned to distinct geographic regions (urban / rural), spatial scales (building / city), and data modalities (remote sensing / trajectory / POI). 
Its sparse activation mechanism substantially reduces computational overhead in processing global-scale geospatial big data, while enabling incremental learning and multi-task optimization. 
Additionally, the multi-agent collaborative framework~\cite{xie2024large,tran2025multiagent,zheng2025graphgeo,han2025swarm} facilitates the modeling of distributed geospatial analysis scenarios, with individual agents handling heterogeneous tasks (traffic analysis / POI mining / image interpretation) and enabling cross-regional knowledge transfer through communication protocols. 
Such a framework is especially suited to addressing the continuous learning requirements inherent in dynamic geographic environments characterized by phenomena such as urban expansion and land use change.

\textbf{Fairness, Equity and Ethics.}
As Geospatial Representation Learning (GRL) systems increasingly inform public policy and resource allocation, the convergence of Fairness, Equity and Ethics becomes paramount. Existing models are frequently trained on data-rich regions (e.g., OpenStreetMap data which is denser in the Global North~\cite{herfort2021evolution}), creating a risk of "representation inequality" where the resulting embeddings perform poorly for under-resourced regions in the Global South. 
Future directions necessitate prioritizing the development of robust transfer learning and few-shot techniques that can generate high-quality representations despite data scarcity~\cite{prabowo2024traffic}. 
Furthermore, to mitigate ethical risks regarding data privacy and the reinforcement of historical prejudices, the academia must advance privacy-preserving frameworks such as Federated Learning~\cite{kammuller2024formalizing} and Differential Privacy~\cite{zhang2023trajectory,jin2022survey}, alongside debiasing techniques like Adversarial Disentanglement~\cite{fan2022debiasing,madras2018learning,zhao2025fairdrl}.
These mechanisms are critical to prevent the ``digitization of segregation'' by ensuring that learned representations do not encode systemic biases, such as redlining, or compromise individual anonymity in automated governance systems.



\vspace{-0.5em}
\section{Conclusion}
\label{sec:conclusion}
In conclusion, this survey highlights the critical role of deep learning in advancing geospatial representation learning. We provide a systematic and detailed overview of modern frameworks leveraging deep neural networks for this purpose, introducing a novel taxonomy organized along three methodological dimensions: modality, coverage, and downstream tasks. Representative studies are systematically reviewed according to this taxonomy, followed by a discussion of current limitations and promising future research directions in the LLM and foundation model era.










\bibliographystyle{elsarticle-harv} 
\bibliography{multimodal_survey_ref}





\end{document}